\documentclass[11pt,parskip=half,DIV=12,final]{scrartcl}
\usepackage[utf8]{inputenc}
\usepackage{amsmath}
\usepackage{amssymb}    % Alex
\usepackage{amsthm}     % Alex
\usepackage[affil-it]{authblk} % affiliations for authors on title page
\usepackage{mathtools}  % Alex
\usepackage{csquotes}   % Alex
\usepackage{tikz}       % Alex
\usepackage{siunitx}
\usepackage{hyperref}
\usepackage{cleveref}
\usepackage{xspace}
\usepackage{xifthen}
\usepackage{nicefrac}
\usepackage{caption}
\usepackage{subcaption}
\usepackage{marvosym}
\usepackage{booktabs}
\usepackage{float}
\usepackage{xcolor}
\usepackage{lineno}
\usepackage[boxed,noend,linesnumbered]{algorithm2e}
\usepackage[textsize=tiny,obeyFinal]{todonotes}
\usepackage[backend=biber,style=numeric,sorting=none,isbn=false,url=false,eprint=false,citestyle=numeric-comp,backref=false,giveninits]{biblatex}
\usepackage{pgfplots}
\pgfplotsset{compat=1.17}

\pgfmathdeclarefunction{gauss}{2}{%
  \pgfmathparse{1/(#2*sqrt(2*pi))*exp(-((x-#1)^2)/(2*#2^2))}%
}

\AtEveryBibitem{\clearfield{month}} % make sure that only year of publication appear in bibliography
\AtEveryBibitem{\clearfield{day}}
\bibliography{references}

\graphicspath{{images/}}

\newcommand{\BFS}{\emph{BFS}}
\newcommand{\ite}{i.\,e.\xspace}
\newcommand{\eg}{e.\,g.\xspace}
\newcommand{\cf}{cf.\xspace}
\newcommand{\etal}{et\,al.\xspace}
\newcommand{\ma}{^\mathrm{(max)}}
\newcommand{\ee}{\mathrm{e\rightarrow{}e}}
\newcommand{\ei}{\mathrm{e\rightarrow{}i}}
\newcommand{\ie}{\mathrm{i\rightarrow{}e}}
\newcommand{\startnode}{\ensuremath{s}\xspace}
\newcommand{\targetnode}{\ensuremath{t}\xspace}
\newcommand{\mean}[1]{\mathop{\operatorname{mean}}\left(#1\right)}
\newcommand{\At}{\ensuremath{\mathcal{A}_{t}}}
\newcommand\redactionmode{2}

\makeatletter
\newcommand{\@minipagerestore}{\setlength{\parskip}{\medskipamount}}
\makeatother

\newcommand{\redact}[2][]{%
    \ifcase\redactionmode%
        \ifthenelse{\isempty{#1}}%
            {%
              {\footnotesize\color{black!40}#2}%
            }%
            {%
                \ifvmode%
                  \begin{minipage}{0.3\textwidth}#1\end{minipage}%
                  \hfill%
                  \begin{minipage}{0.6\textwidth}\footnotesize\color{black!40}#2\end{minipage}%
                \else%
                  \par%
                  \begin{minipage}{0.3\textwidth}#1\end{minipage}%
                  \hfill%
                  \begin{minipage}{0.6\textwidth}\footnotesize\color{black!40}#2\end{minipage}%
                  \par%
                \fi
            }
    \or%
        #1%
    \or%
        #2%
    \fi%
}
\Crefname{algocf}{Algorithm}{Algorithms}
\DeclareSIUnit\seema{\ensuremath{s_\ee\ma}}

\newcommand{\V}{\mathcal{V}}
\newcommand{\E}{\mathcal{E}}

\newcolumntype{d}[1]{D{.}{.}{#1}}

\title{Can the brain use waves to solve planning problems?}

\author[1,2]{Henry~Powell}
\author[1]{Mathias~Winkel}
\author[1]{Alexander~V.~Hopp}
\author[1]{Helmut~Linde}
\affil[1]{Merck~KGaA, Darmstadt, Germany}
\affil[2]{University~of~Glasgow, Scotland}
\date{\today}

\begin{document}
\maketitle
\begin{abstract}
    A variety of behaviors like spatial navigation or bodily motion can be formulated as graph traversal problems through cognitive maps.
    We present a neural network model which can solve such tasks and is compatible with a broad range of empirical findings about the mammalian neocortex and hippocampus.
    The neurons and synaptic connections in the model represent structures that can result from self-organization into a cognitive map via Hebbian learning, i.e. into a graph in which each neuron represents a point of some abstract task-relevant manifold and the recurrent connections encode a distance metric on the manifold.
    Graph traversal problems are solved by wave-like activation patterns which travel through the recurrent network and guide a localized peak of activity onto a path from some starting position to a target state.
\end{abstract}

\clearpage
\listoftodos

\clearpage
\tableofcontents

% =====================================================================
\section{Introduction}
% =====================================================================

\redact[
    Building a bridge between structure and function of neural networks is an ambition at the heart of neuroscience.
    Historically, the first models studied were simplistic artificial neurons arranged in a feed-forward architecture.
    Such models are still widely applied today -- forming the conceptual basis for Deep Learning. 
    They have shaped our intuition of neurons as \enquote{feature detectors} which fire when a certain approximate configuration of input signals is present, and which aggregate simple features to more and more complex ones layer by layer. 
    Yet in the brain, the vast majority of neural connections  is recurrent, and although several possible explanations of their function have been proposed \cite{singer_does_2016,miller_cortical_2013, kveraga_top-down_2007}, their computational purpose is still little understood \cite{douglas_recurrent_2007}.
]{
    Understanding the computational principles of the human brain is one of the most ambitious goals of neuroscience, and one which promises vast intellectual and practical benefits. Yet in the face of its enormous complexity, even after decades of intense research, the understanding of the brain's algorithms remains very vague, at best.

    Some level of insight stems from the thorough analysis of neural feed-forward architectures.
    There, a neuron is usually considered to be an electrical component which computes an output by applying a non-linear function on some weighted sum of its synaptic inputs and transmits the result to a next higher layer of neurons.

    This simplistic but effective model has been exploited in many technical applications in the form of (deep) artificial neural networks. Their neurons are typically organized in layers, each of which sends its output signals only to the next higher layer. Such feed-forward processing has shaped our intuition of neurons as \enquote{feature detectors} which fire when a certain approximate configuration of input signals is present, and which aggregate simple features to more and more complex ones layer by layer.

    In the brain, though, the overwhelming majority of connections between neurons are recurrent, \ite they connect neurons within the same cortical area or transmit information from higher areas back to lower ones.
    For example, in the visual cortex, synapses from the lateral geniculate nucleus of the thalamus, \ite the feed-forward connections, make up only \SIrange[range-phrase=--]{5}{10}{\percent} of the excitatory synapses in their target layer 4 of V1 in cats and monkeys \cite{douglas_recurrent_2007}.
    The understanding of neurons as \enquote{feature detectors} can therefore only represent a small fragment of the over-all picture.

    Several possible explanations of the function of these recurrent connections have been proposed. For example, it has been suggested that neural activity follows almost chaotic trajectories in an extremely high-dimensional state space while the dynamics are still sensitive enough to be influenced by the relatively small share of feed-forward connections \cite{singer_does_2016}. In this conceptual framework,  attention signals, past memories, and sensory input are merged in order to guide the system towards well-separated, lower-dimensional subspaces which represent certain states of perception. It is also hypothesized that top-down projections from higher cortical areas transmit predictions or expectations to influence how the lower areas interpret the incoming sensory data \cite{miller_cortical_2013, kveraga_top-down_2007}. Such predictions are thought to play a role in noise-reduction and signal-restoration or to direct attention bottom-up to features which deviate from the prediction and thus require some executive reaction. Nevertheless, the full computational purpose of the recurrent connections is still little understood \cite{douglas_recurrent_2007}.
}

In the present paper, we propose a new algorithmic role which recurrent neural connections might play, namely as a computational substrate to solve graph traversal problems.
We argue that many cognitive tasks like navigation or motion planning can be framed as finding a path from a starting position to some target position in a space of possible states.
The possible states may be encoded by neurons via their \enquote{feature-detector property}. Allowed transitions between nearby states would then be encoded in recurrent connections, which can form naturally via Hebbian learning since the feature detectors' receptive fields overlap.
They may eventually form a \enquote{map} of some external system.
Activation propagating through the network can then be used to find a short path through this map.
In effect, the neural dynamics then implement an algorithm similar to Breadth-First Search on a graph.

\redact{
    The remainder of the paper is organized as follows:
    In \Cref{sec:model}, we give a conceptual overview, describe the technical details of the proposed model and show some simulation results for an exemplary numerical implementation of the model.
    We then review empiric support for some components of the model in \Cref{sec:empiric_evidence}.
    Limitations, implications and ideas for further development are discussed in \Cref{sec:discussion}.
    The more technical details related to general graph theory and to the numerical implementation can be found in \Cref{sec:methods}.
}

% =====================================================================
\section{Proposed Model} \label{sec:model}
% =====================================================================

\subsection{A Network of Neurons that Represents a Manifold of Stimuli}\label{sec:model:manifold}

\begin{figure}[tb]
    \hfill
    \begin{subfigure}[t]{0.32\textwidth}
        \centering
        \includegraphics[width=0.8\textwidth]{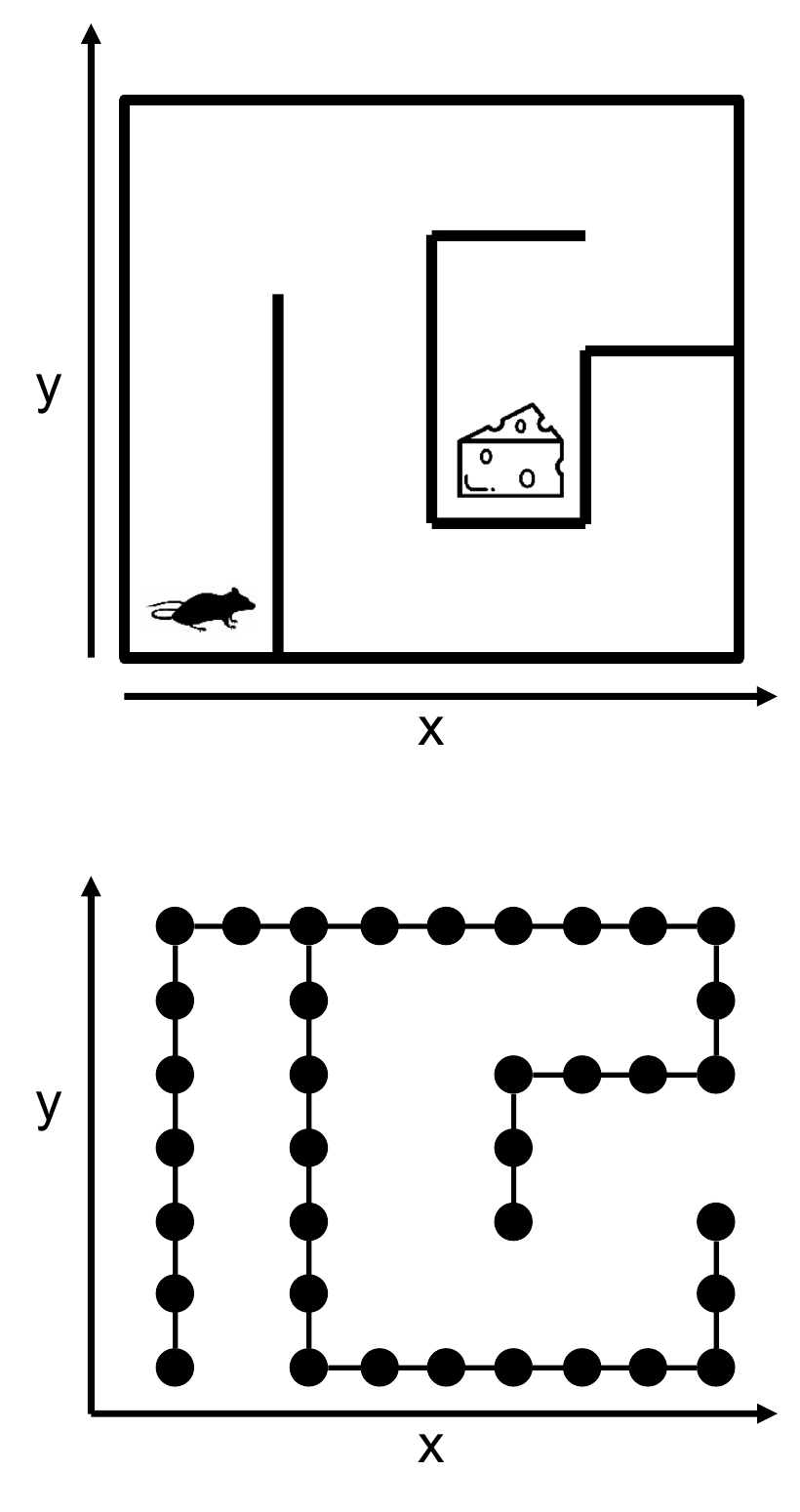}
        \caption{%
            Approximate positions in the maze are encoded in single neurons.
            \redact{Overlapping receptive fields lead to recurrent connections which resemble the structure of the maze.}
            The planning problem is to find a way through the maze given the current position of the cheese and the mouse.}
        \label{fig:examples:maze}
    \end{subfigure}
    \hfill
    \begin{subfigure}[t]{0.32\textwidth}
        \centering
        \includegraphics[width=0.8\textwidth]{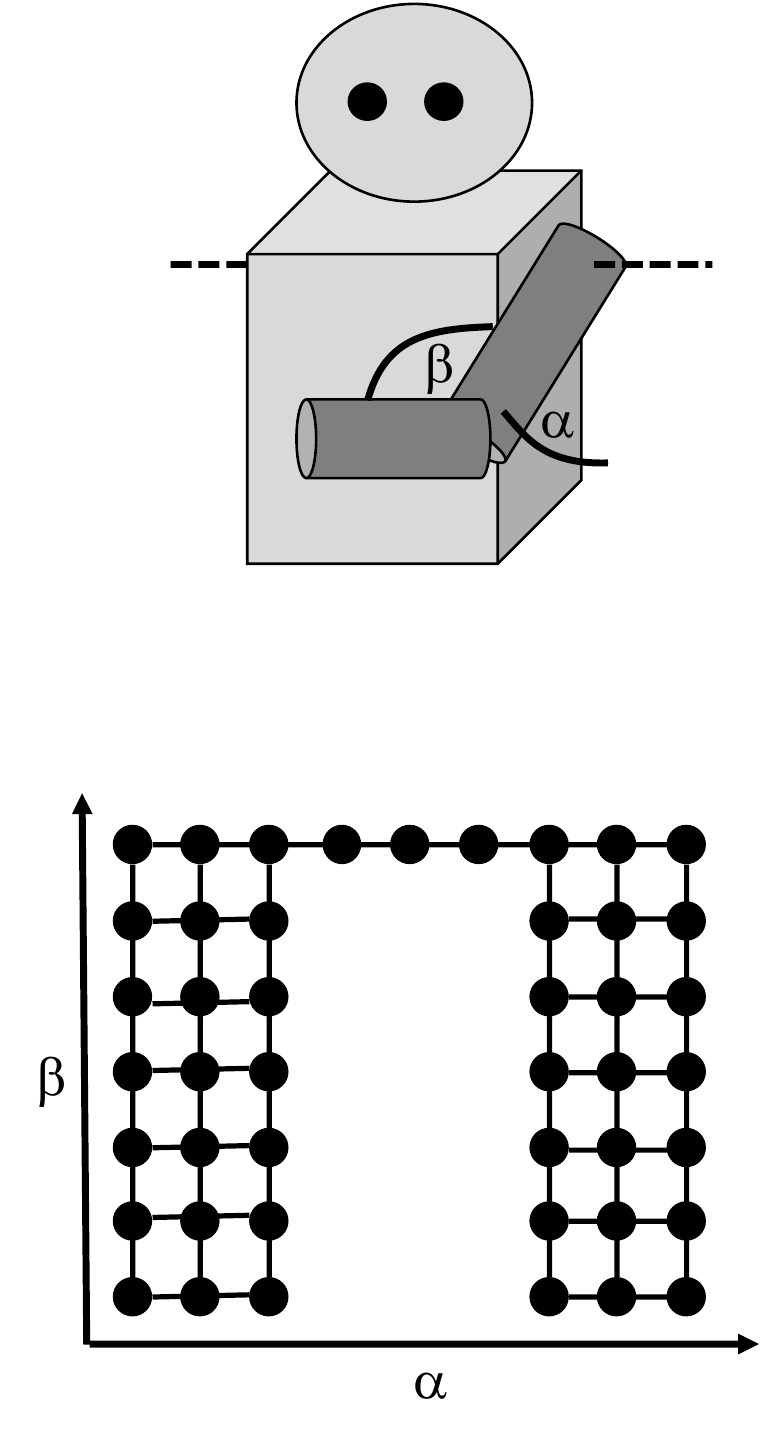}
        \caption{%
            Approximate positions of the \enquote{arm} are encoded in single neurons.
            Physically impossible positions \redact{where the \enquote{arm} intersects with the \enquote{body}} are not encoded at all \redact{(because they have never been observed by the neural network)} giving rise to the gap in the center of the cognitive map.
            An example planning problem is to move the \enquote{hand} from behind the body to a position in front of the body without collision.}
        \label{fig:examples:robot}
    \end{subfigure}
    \hfill
    \begin{subfigure}[t]{0.32\textwidth}
        \centering
        \includegraphics[width=0.8\textwidth]{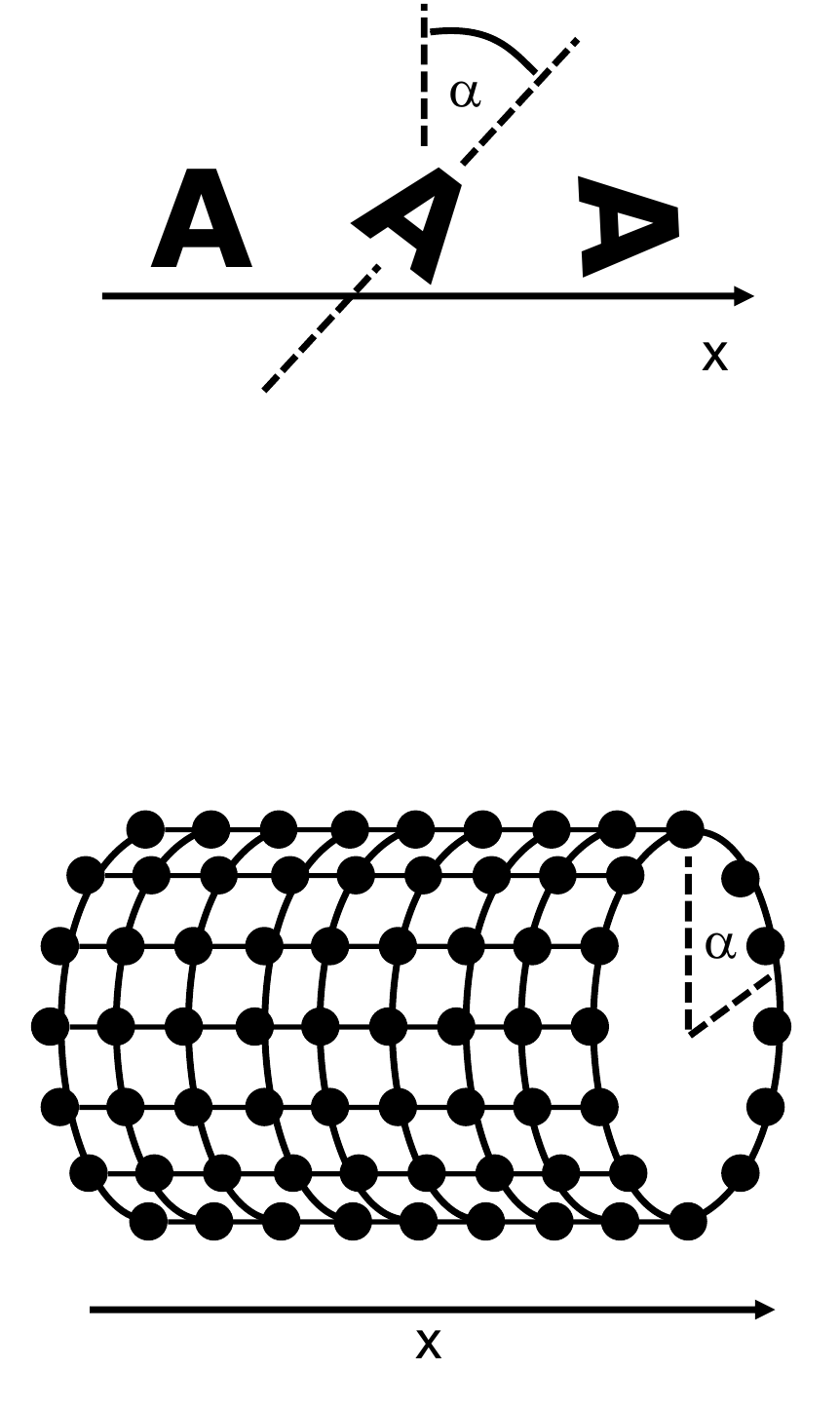}
        \caption{%
            The visual stimulus is always the letter \enquote{A}, but at different $x$-positions and tilted at different angles $\alpha$.
            \redact{Due to the periodicity of the stimulus under change of $\alpha$, the resulting cognitive map has the topology of a cylinder.}
            An example planning problem in this case is the decision whether the \enquote{A} has to be moved/tilted to the left or to the right to convert it from some given position to another one.
        }
        \label{fig:examples:A}
    \end{subfigure}
    \caption{Three examples of stimuli-generating processes and recurrent neural networks representing the corresponding manifold of stimuli.}
    \label{fig:examples}
\end{figure}

We consider a neural network which is exposed to some external stimuli-generating process under the assumption that the possible stimuli can be organized in some continuous manifold\redact{\footnote{In mathematics, a manifold is a topological space which has the structure of a Euclidean space locally at each point. In contrast to a (globally) Euclidean space, manifolds can be topologically diverse and -- when endowed with a Riemannian metric -- curved.
For example, a saddle-shaped hyperbolic plane, a sphere or a torus are manifolds.}} in the sense that similar stimuli are located close to each other on this manifold.
For example, in the case of a mouse running through a maze all possible perceptions can be associated with a particular position in a two-dimensional map, and neighboring positions will generate similar perceptions, see \Cref{fig:examples:maze}.

Proprioception, \ite the sense of location of body parts, can also be a source of stimuli.
For example, for a simplified arm with two degrees of freedom every possible position of the arm corresponds to one specific stimulus, \cf \Cref{fig:examples:robot}.
All possible stimuli combined give rise to a two-dimensional manifold.
The example also shows that the manifold will usually be restricted since not every conceivable combination of two joint angles might be a physically viable position for the arm.

The manifold of potential stimuli needs not necessarily be embedded in a flat Euclidean space as in the case of the maze.
For example, if the stimuli are two-dimensional figures which can be shifted horizontally or rotated on a screen, the corresponding manifold is two-dimensional (one translational parameter plus one for the rotation angle) but it is not isomorphic to a flat plane since a change of the rotation angle by $2\pi$ maps the figure onto itself again, see \Cref{fig:examples:A}.

We assume that such manifolds of stimuli are approximated by the connectivity structure of a neural network which forms via a learning process.
The result is a neural structure which we call a \emph{cognitive map}.
The defining property of a cognitive map is that is has a neural encoding for every possible stimulus and that two similar stimuli, \ite stimuli which are close to each other in the manifold of stimuli, are represented by similar encodings, \ite encodings which are close to each other in the cognitive map\footnote{Of course, we do not imply that two neurons which are close to each other in the connectivity structure are also close to each other with respect to their physical location in the neural tissue.}.
\redact{
    There is considerable evidence, which we review in \Cref{sec:cognitive_maps}, that such cognitive maps are implemented by the brain, but the details of the encoding of stimuli remain mostly unclear.
}

\begin{figure}[tb]
    \centering
    \includegraphics[width=0.7\textwidth]{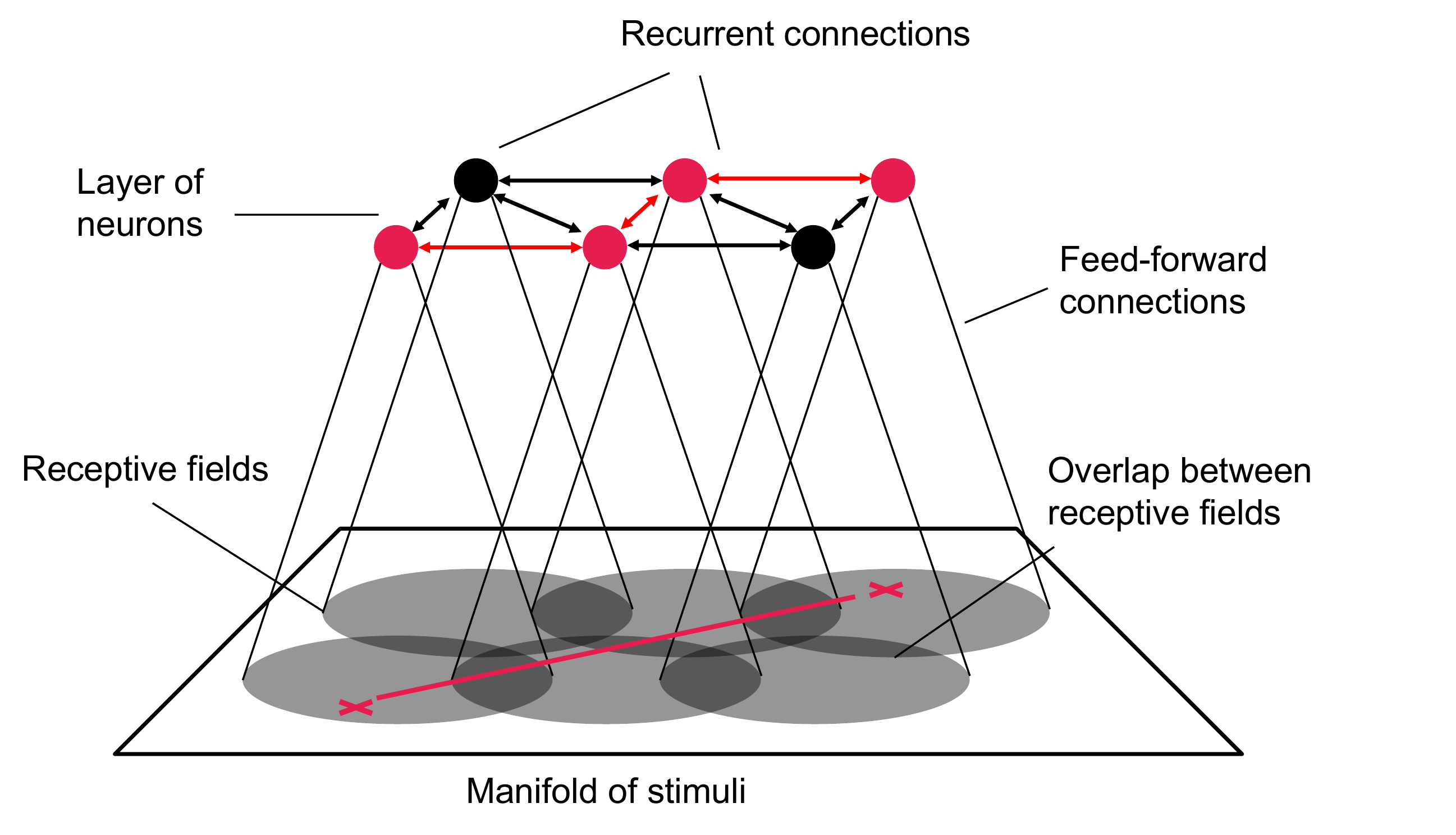}
    \caption{%
        In the model, the recurrent connections within a single layer of neurons approximate the topology of the manifold of stimuli.
        During the learning process, the strongest recurrent connections are formed between neurons with overlapping receptive fields. %, thus preferentially connecting neurons which encode neighboring positions in the manifold.
        The problem of finding a route through the manifold (red line) is thus approximated by the problem of finding a path through the graph of recurrent neural connections (red path).
        }
    \label{fig:recurrent}
\end{figure}

For the model, we make a very simplistic choice and assume a single-neuron encoding, \ite the manifold of stimuli is covered by the receptive fields of individual neurons. Each such receptive field is a small localized area in the manifold and two neighboring receptive fields may overlap, see \Cref{fig:recurrent}.
Such an encoding is a typical outcome for a single layer of neurons which are trained in a competitive Hebbian learning process \cite{rumelhart_feature_1985}. %, \ite where only the most highly activated neurons adapt their synaptic weights to the input signal in each learning step.
\redact{Examples for such competitive learning algorithms are Kohonen Maps \cite{kohonen_self-organized_1982}, (Growing) Neural Gas \cite{martinetz_topology_1994} or variants of sparse coding dictionary learning \cite{Elad:2010}.}

The key idea of the model is that solving a problem that can be formulated as a planning problem in the manifold of stimuli, can be solved by a planning problem in a corresponding cognitive map.
To this end, it is not enough to consider the cognitive map as a set of individual points, but its topology must be known as well.
This topological information will be encoded in the recurrent connections of the neural network.
It seems natural that a neural network could learn this topology via Hebbian learning:
Two neurons with close-by receptive fields in the manifold will be excited simultaneously relatively often because their receptive fields overlap. Consequently, recurrent connections within the cognitive map will be strengthened between such neurons and the topology of the neural network will approximate the topology of the manifold, see \Cref{fig:recurrent}. This idea has been explored in more detail by Curto and Itskov in \cite{curto_cell_2008}.

\subsection{Dynamics Required for Solving Planning Problems}\label{sec:model:dynamics}

Having set up a network that represents a manifold of stimuli, we need to endow this network of feed-forward and recurrent connections with dynamics. % such that it solves a planning problem.
We do so by imposing two interacting mechanisms.

First, the neurons in the network should exhibit continuous attractor dynamics \cite{rolls_attractor_2010}: 
If a \enquote{clique} of a few tightly connected neurons are activated by a stimulus via the corresponding feed-forward pass, they keep activating each other while inhibiting their wider neighborhood.
The result is a self-sustained, localized neural activity surrounded by a \enquote{trench of inhibition}.
In the model, this encodes the as-is situation or the starting position for the planning problem.
Such a state is called an \enquote{attractor} since it is stable under small perturbations of the dynamics, and it is part of a continuous landscape of attractors with different locations across the network. 
\redact{
    For a recent review of attractor networks, the reader is referred to \cite{rolls_attractor_2010}.
}

Second, the neural network should allow for wave-like expansion of activity.
If a small number of close-by neurons are activated by some hypothetical executive brain function (\ite not via the feed-forward pass), they activate their neighbors, which in turn activate theirs, and so on. The result is a wave-like front of activity propagating through the recurrent network. The neurons which have been activated first encode the to-be state or the end position of the planning problem.

\begin{figure}[tb]
    \centering
    \includegraphics[width=0.7\textwidth]{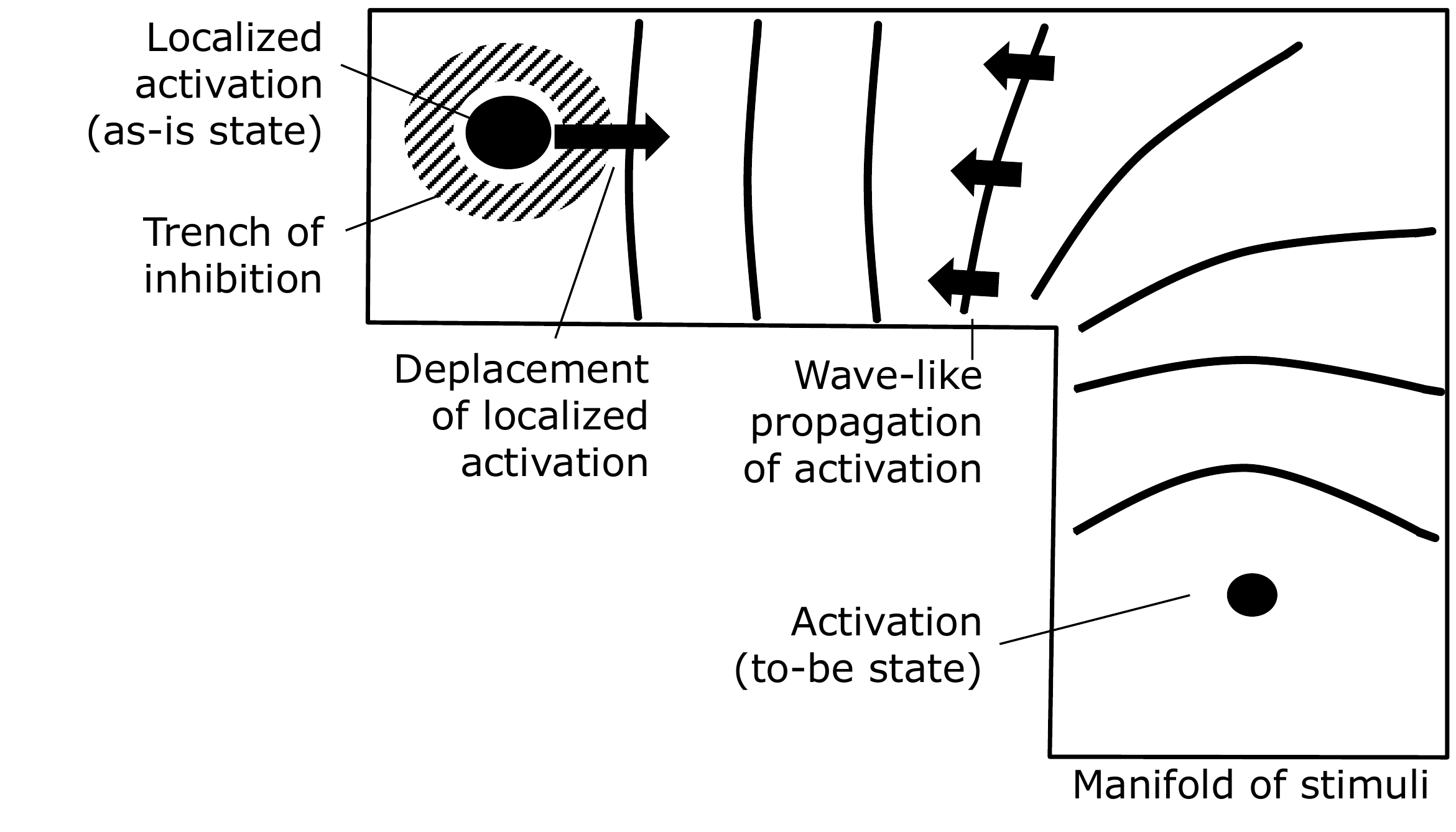}
    \caption{%
        The as-is state of the system is encoded in a stable, localized, and self-sustained peak of activity surrounded by a \enquote{trench} of inhibition (top left corner). A planning process is started by stimulating the neurons which encode the to-be position (bottom right corner). The resulting waves of activity travel through the network and interact with the localized peak. Each incoming wave front shifts the peak slightly towards its direction of origin. Note that, for reasons of simplicity, we did not draw the neural network in this figure but only the manifold which it approximates.
        }
    \label{fig:waves}
\end{figure}

The key to solving a planning problem is in the interaction between the two types of dynamics, namely in what happens when the expanding wave front hits the stationary peak of activity. 
%Remember that the peak has created a \enquote{trench of inhibition} around itself which stabilizes its position. 
On the side where the wave is approaching it, the \enquote{trench of inhibition} surrounding the peak is in part neutralized by the additional excitatory activation from the wave. 
Consequently, the containment of the activity peak is somewhat \enquote{softer} on the side where the wave hit it and it may move a step towards the direction of the incoming wave. 
This process repeats, leading to a small change of position with every incoming wave front. The localized peak of excitation will follow the wave fronts back to their source, thus moving along a route through the manifold from start to end position, see \Cref{fig:waves}.

The two types of dynamics described above are seemingly contradictory, since the first one restricts the system to localized activity, while the second one permits a wave-like propagation of activity throughout the system.
To resolve the conflict in numerical simulations, we have split the dynamics into a \emph{continuous attractor layer} and a \emph{wave propagation layer}, which are responsible for different aspects of the system's dynamical behaviour. We discuss the concepts of a  numerical implementation in \Cref{sec:model:implementation} and ideas for a biologically more plausible implementation in \Cref{sec:discussion}.

\subsection{Connection to Real-Life Cognitive Processes}\label{sec:model:connection}

To make the proposed concept more tangible we present a rough sketch of how it could be embedded in a real-life cognitive process.
As an example, we consider a human grabbing a cup of coffee.
According to our hypothesis, the as-is position of the subject's arm is encoded as a localized peak of activity in the cognitive map encoding the complex manifold of arm positions.
We assume that this encoding works in a bi-directional way, somewhat like the string of a puppet:
When the arm is moved by external forces, the neural representation of its position moves along with it.
On the other hand, if the representation is changed slightly by some cognitive process, then some hypothetical muscle control mechanism attempts to make the arm follow its neural representation and bring the limb and its representation back into congruence.
If now the human subject decides to grab the cup of coffee, some executive brain function constructs a to-be state of holding the cup:
The position of the hand with the fingers around the cup handle is what the person consciously thinks of.
This thought activates the encoding of the to-be state in the cognitive map that represents the manifold of possible arm positions. The activation creates waves of activity propagating through the network, reaching the representation of the as-is state and shifting it slightly towards the to-be state.
The hypothetical muscle control mechanism reacts on this disturbance and performs a motor action to keep the arm and its representation in line. As long as the person implicitly represents the to-be state, the arm \enquote{automatically} performs the complicated sequence of many individual joint movements which is necessary to grab the cup.

This concept can be extended to flexibly consider restrictions that have not been hard-coded in the cognitive map by learning.
For example, in order to grab the cup of coffee, the arm may need to avoid obstacles on the way.
To this end, the hypothetical executive brain function which defines the target state of the hand could also temporarily \enquote{block} certain regions of the cognitive map (\eg via inhibition) which it associates with the discomfort of a collision. 
Those parts of the network which are blocked cannot conduct the \enquote{planning waves} anymore and thus a path around those regions will be found.

\subsection{Implementation in a Numerical Proof-of-Concept}\label{sec:model:implementation}

To substantiate the presented conceptual ideas, we performed numerical experiments using multiple different setups.
In each case, the implementation of the model employs two neural networks that both represent the same manifold of stimuli.

The continuous attractor layer is a sheet of neurons that models the functionality of a network of place cells in the human hippocampus \cite{Okeefe1971, Okeefe1976}.
Each neuron is implemented as a rate-coded cell embedded in its neighborhood via short-range excitatory and long-range inhibitory connections as in \cite{Guanella2007}.
This structure allows the formation of a self-sustaining \enquote{bump} of activity, which can be shifted through the network by external perturbations.
\redact{The bump represents the as-is state of the planning problem, which is to be solved by moving the bump to its target state.}

The wave propagation layer is constructed with an identical number of excitatory and inhibitory Izhikevich neurons \cite{Izhikevich:2003, Izhikevich:2004}, properly connected to allow for stable signal propagation across the manifold of stimuli. The target node is permanently stimulated, causing it to emit waves of activation which travel through the network. 

The interaction between the two layers is modeled in a rather simplistic way. As in \cite{Guanella2007}, a time-dependent direction vector was introduced in the synaptic weight matrix of the continuous attractor layer. It has the effect of shifting the synaptic weights in a particular direction which in turn causes the location of the activation bump in the attractor layer to shift to a neighbouring neuron. The direction vector is updated whenever a wave of activity in the wave propagation layer newly enters the region which corresponds to the bump in the continuous attractor layer. Its direction is set to point from the center of the bump to the center of the overlap area between bump and wave, thus causing a shift of the bump towards the incoming wave fronts. 

For more details on the implementation, see \Cref{sec:methods:implementation}.

\subsection{Results of the Numerical Experiments} \label{sec:model:experiments}

\begin{figure}[tb]
    \includegraphics[width=0.16\textwidth]{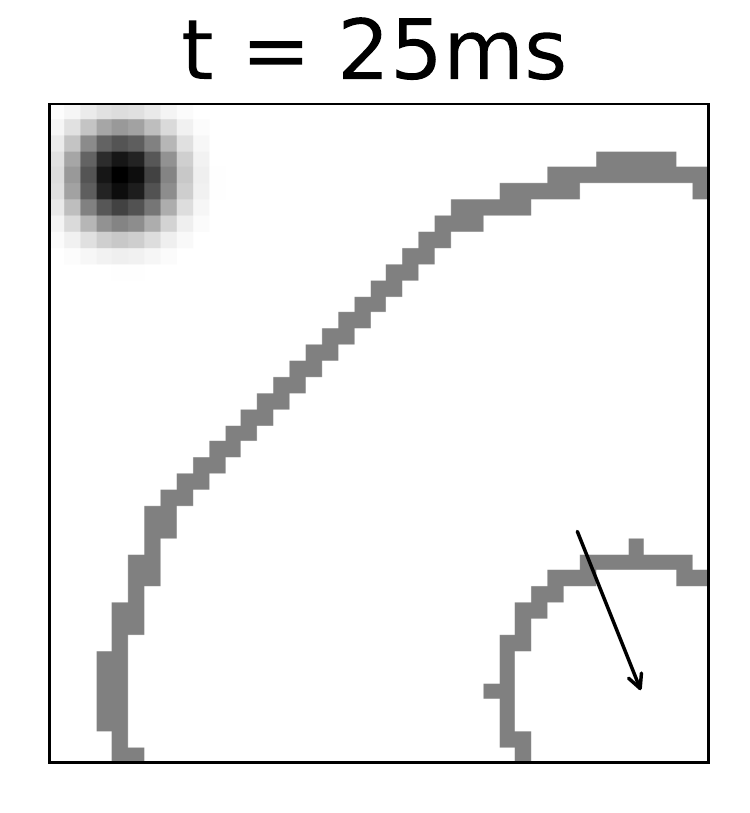}\hfill%
    \includegraphics[width=0.16\textwidth]{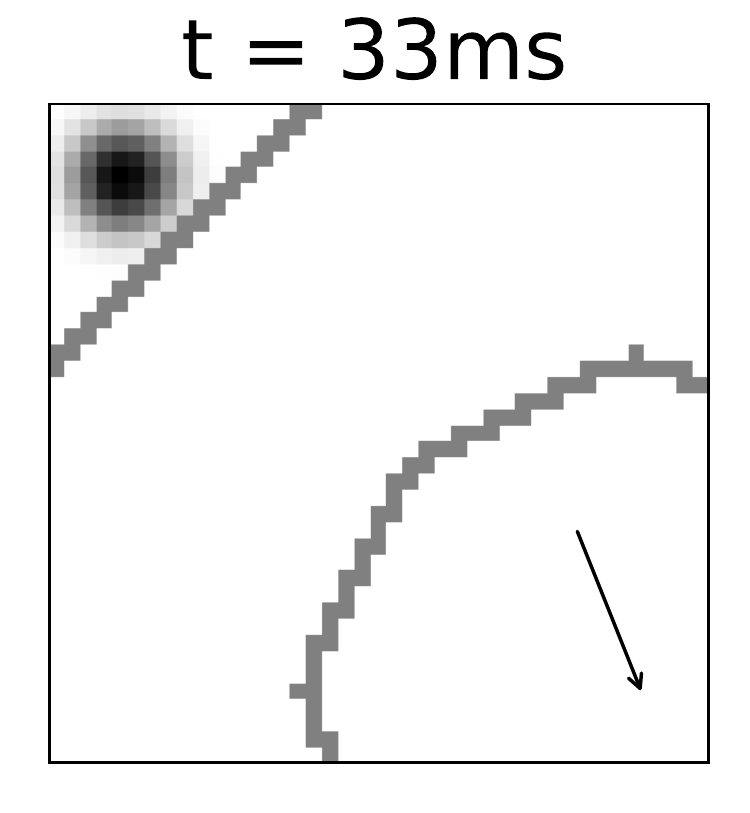}\hfill%
    \includegraphics[width=0.16\textwidth]{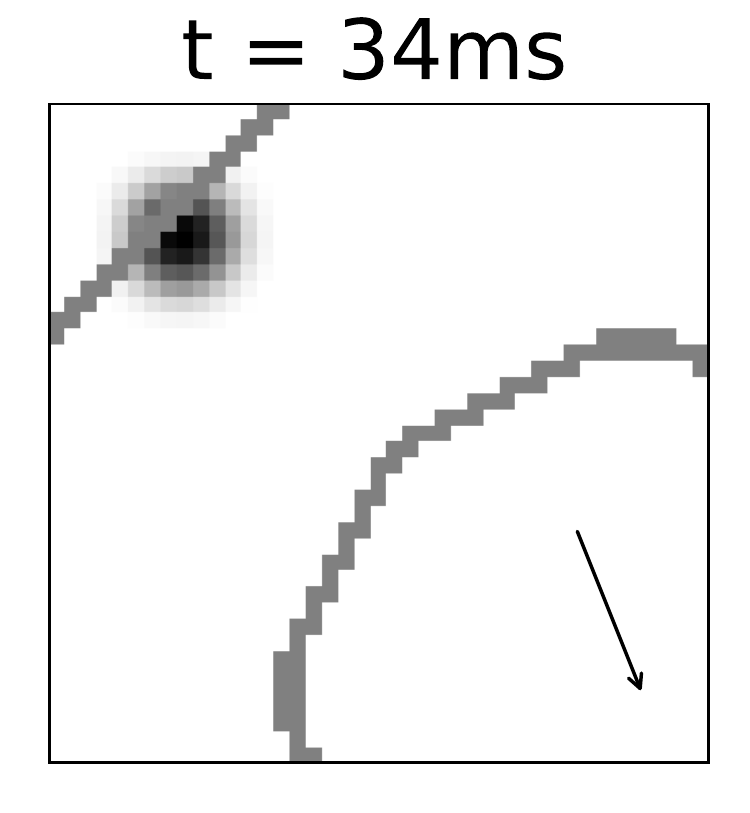}\hfill%
    \includegraphics[width=0.16\textwidth]{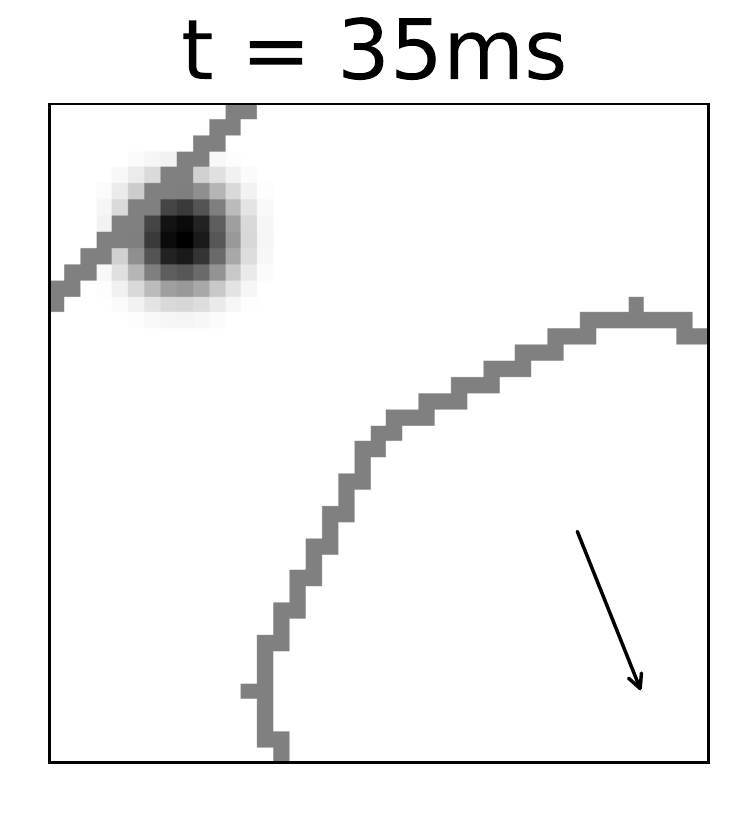}\hfill%
    \includegraphics[width=0.16\textwidth]{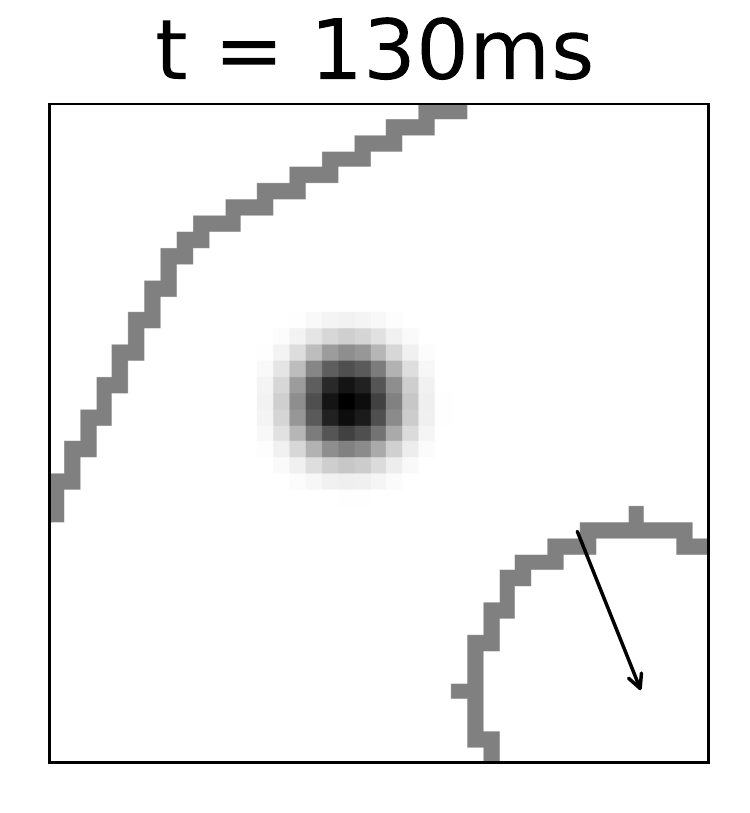}\hfill%
    \includegraphics[width=0.16\textwidth]{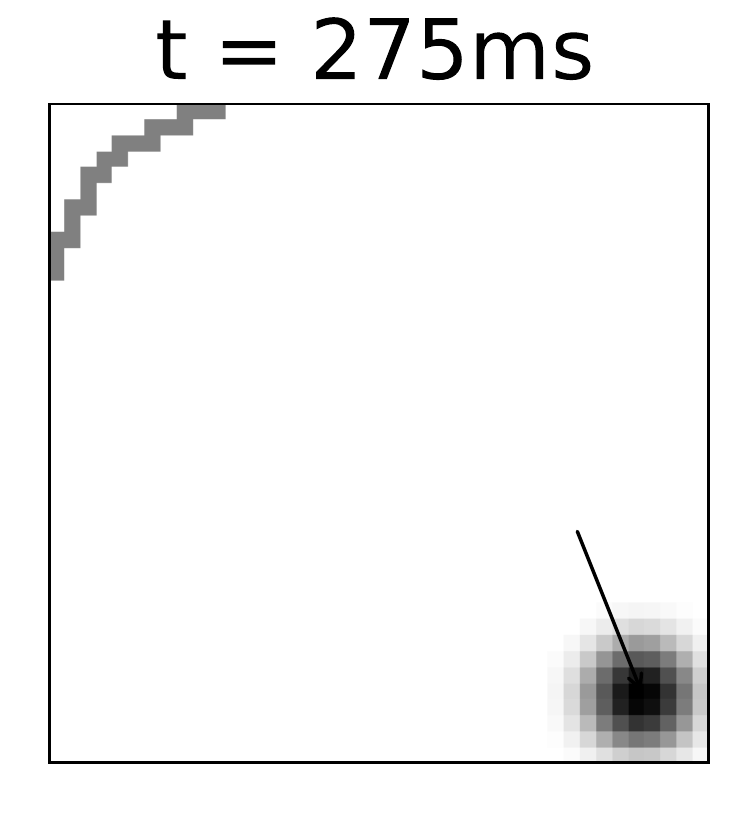}%
    \caption{%
            Activity in the wave propagation layer (greyish lines) and the continuous attractor layer (circular blob-like structure) overlaid on top of each other at different time points during the simulation.
            The position of the external wave propagation layer stimulation (to-be state) is shown with an arrow.
            Starting from an initial position in the top left of the sheet, the activation bump traces back the incoming waves to their source in the bottom right.
        }
    \label{fig:simple_setup}
\end{figure}

In a very simple initial configuration, the path finding algorithm was tested on a fully populated quadratic grid of neurons as described before.
\Cref{fig:simple_setup} shows snapshots of wave activity and continuous attractor position at some representative time points during the simulation.
As expected, stimulation of the wave propagation layer in the lower right of the cognitive map causes the emission of waves, which in turn shift the bump in the continuous attractor layer from its starting position in the upper left towards its target state.

\begin{figure}[tb]
    \centering
    \begin{subfigure}[b]{\textwidth}
        \includegraphics[width=0.16\textwidth]{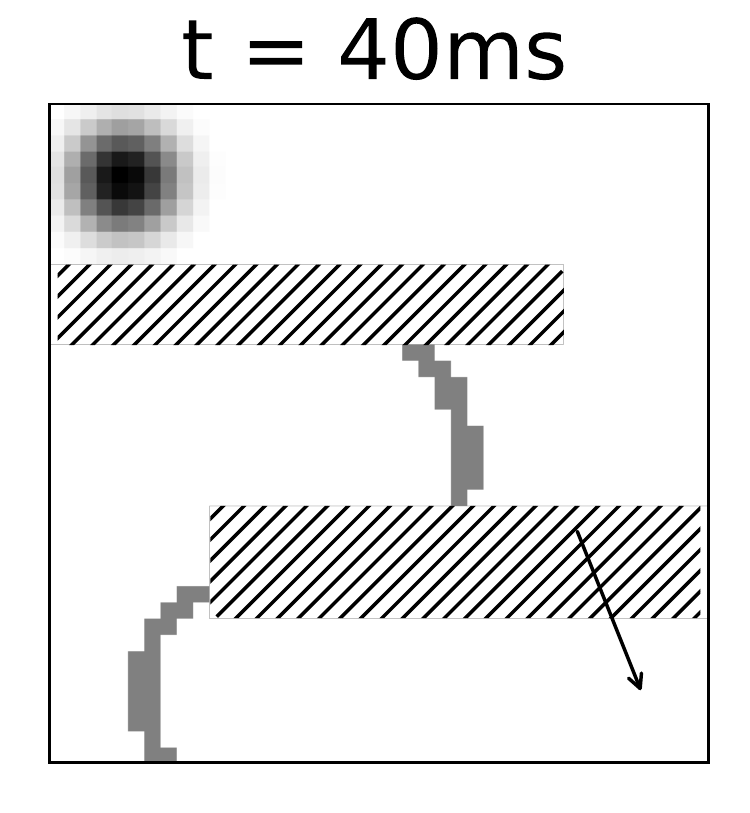}\hfill%
        \includegraphics[width=0.16\textwidth]{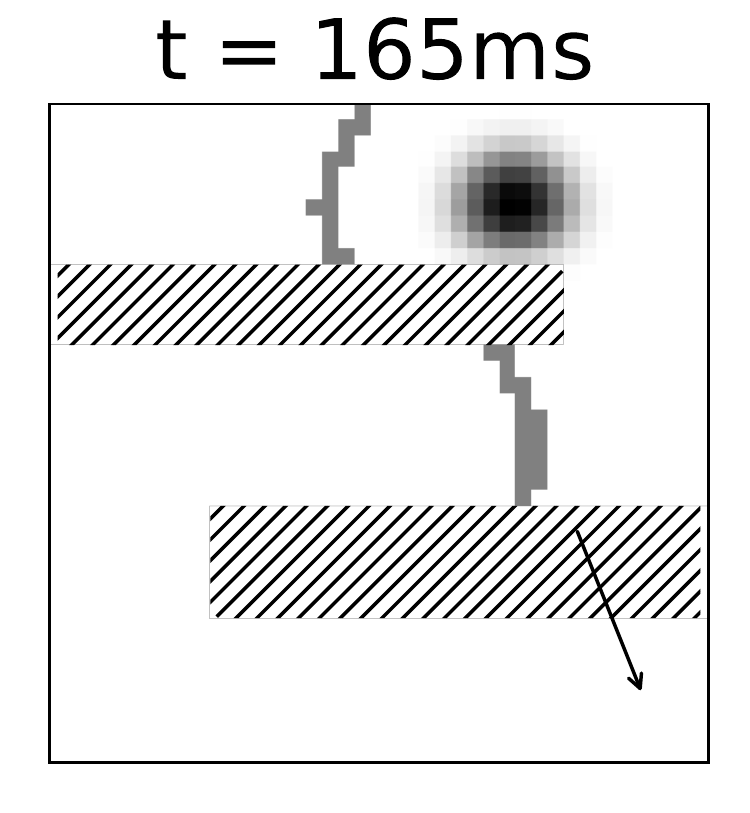}\hfill%
        \includegraphics[width=0.16\textwidth]{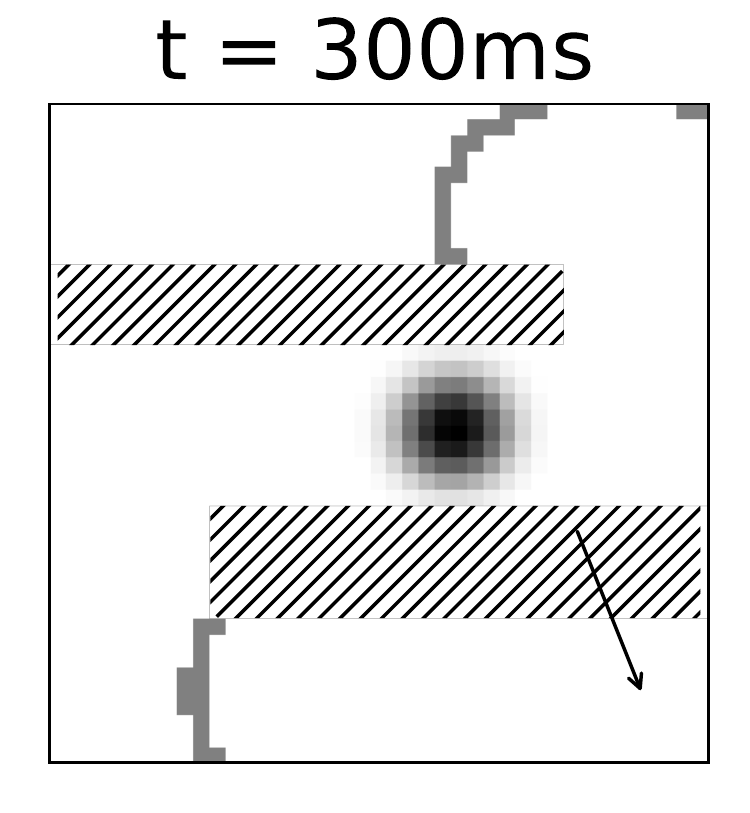}\hfill%
        \includegraphics[width=0.16\textwidth]{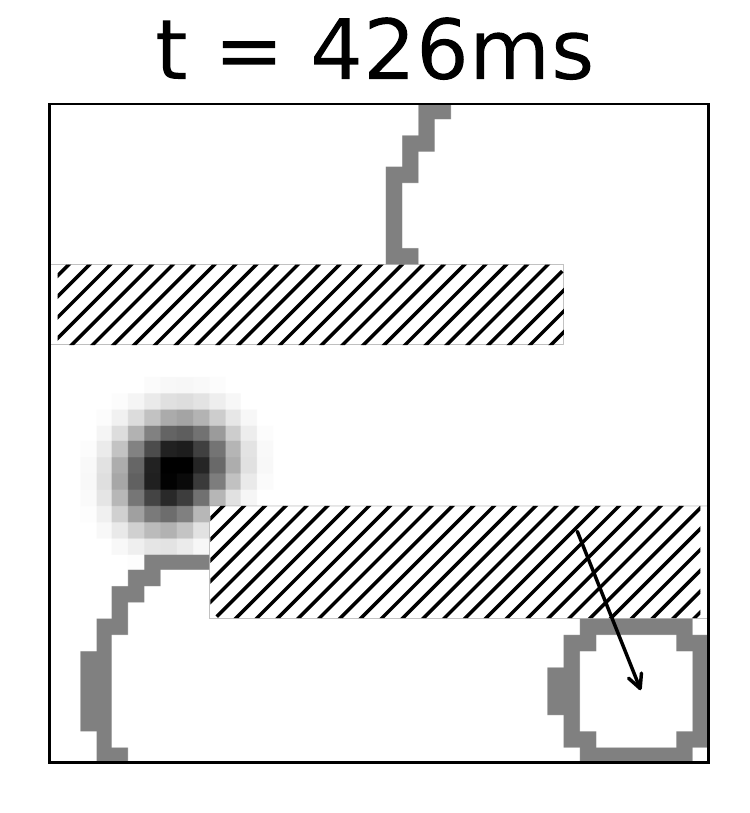}\hfill%
        \includegraphics[width=0.16\textwidth]{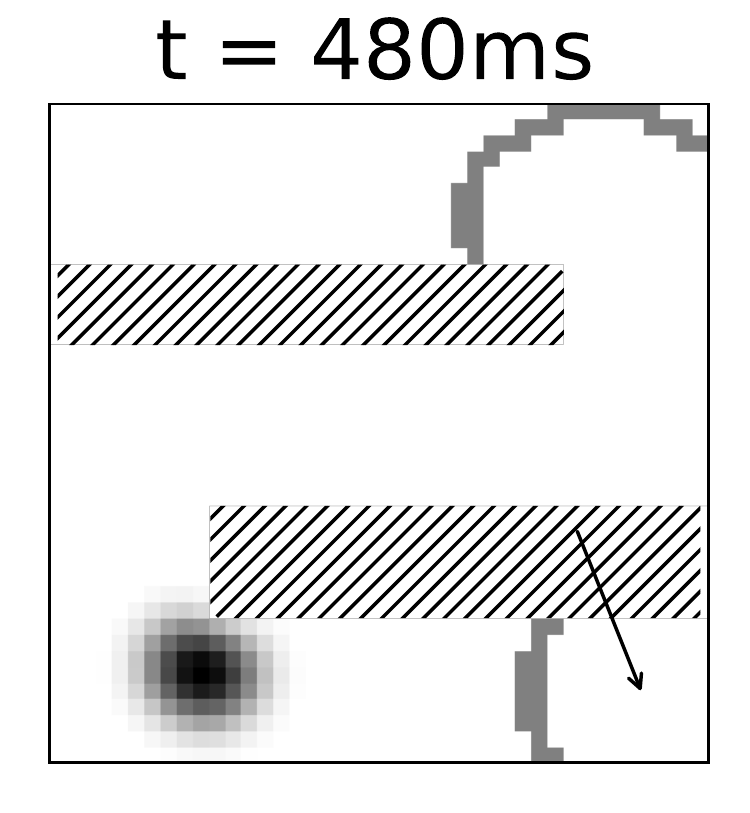}\hfill%
        \includegraphics[width=0.16\textwidth]{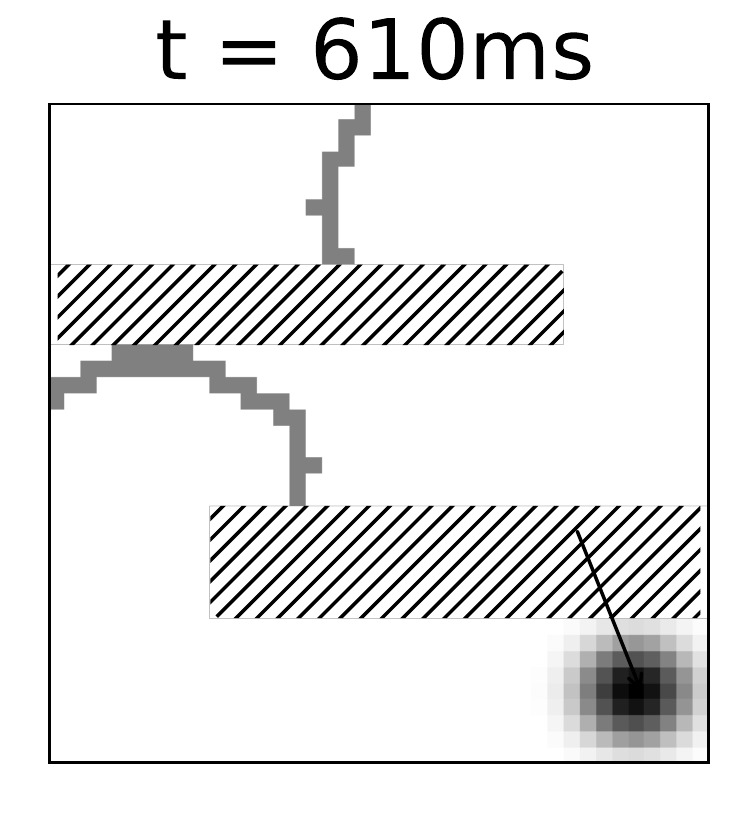}%
        \caption{S Maze}
        \label{fig:mazes:s}
    \end{subfigure}
    \\[.5em]
    \begin{subfigure}[b]{\textwidth}
        \includegraphics[width=0.16\textwidth]{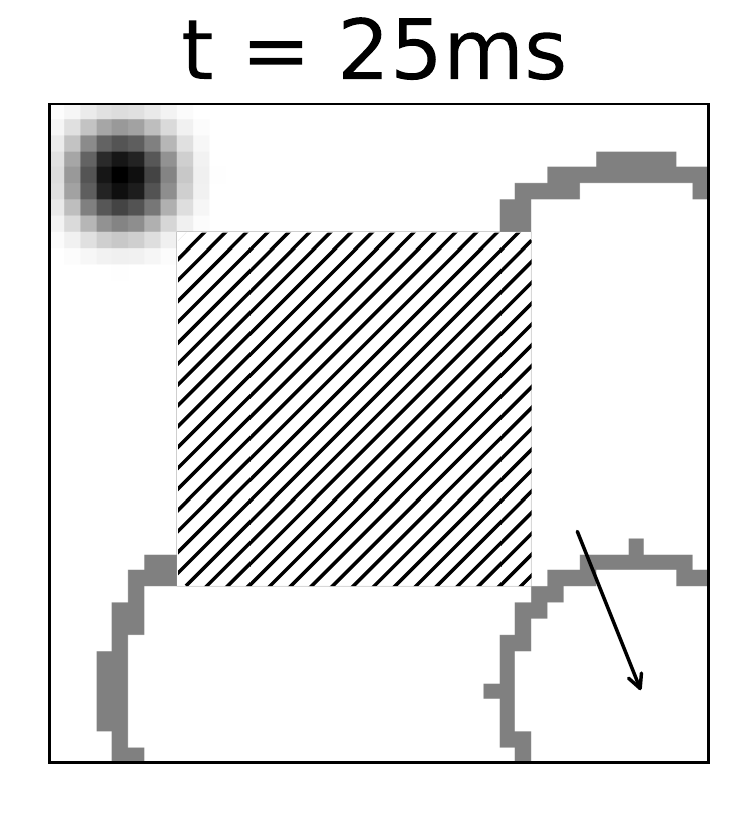}\hfill%
        \includegraphics[width=0.16\textwidth]{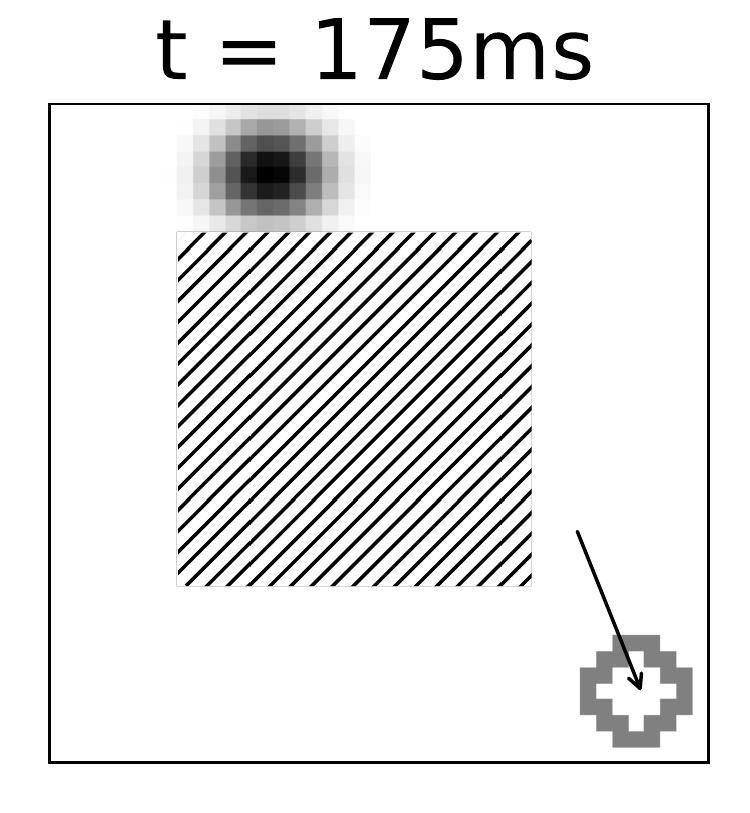}\hfill%
        \includegraphics[width=0.16\textwidth]{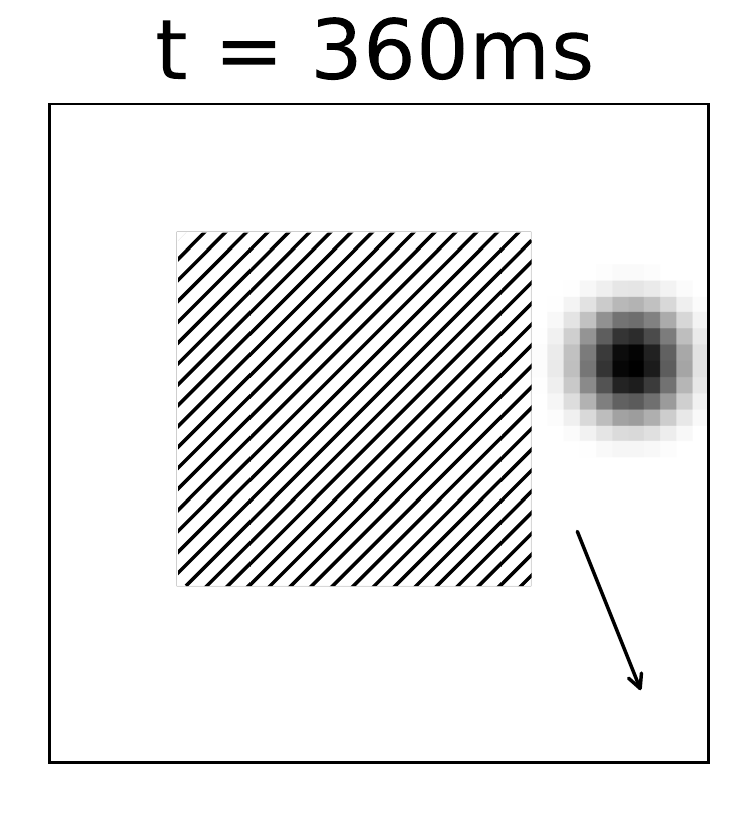}\hfill%
        \includegraphics[width=0.16\textwidth]{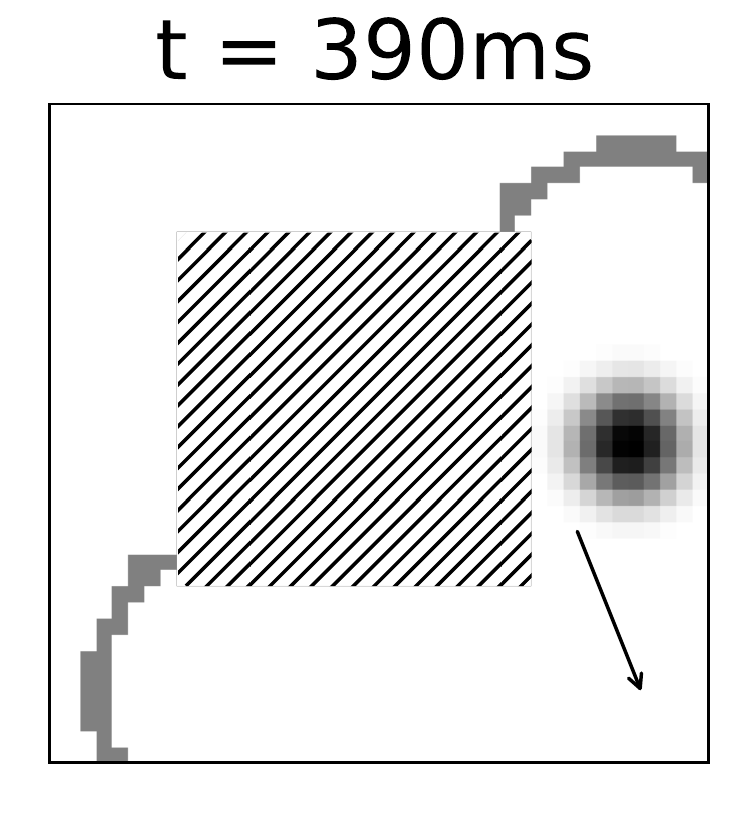}\hfill%
        \includegraphics[width=0.16\textwidth]{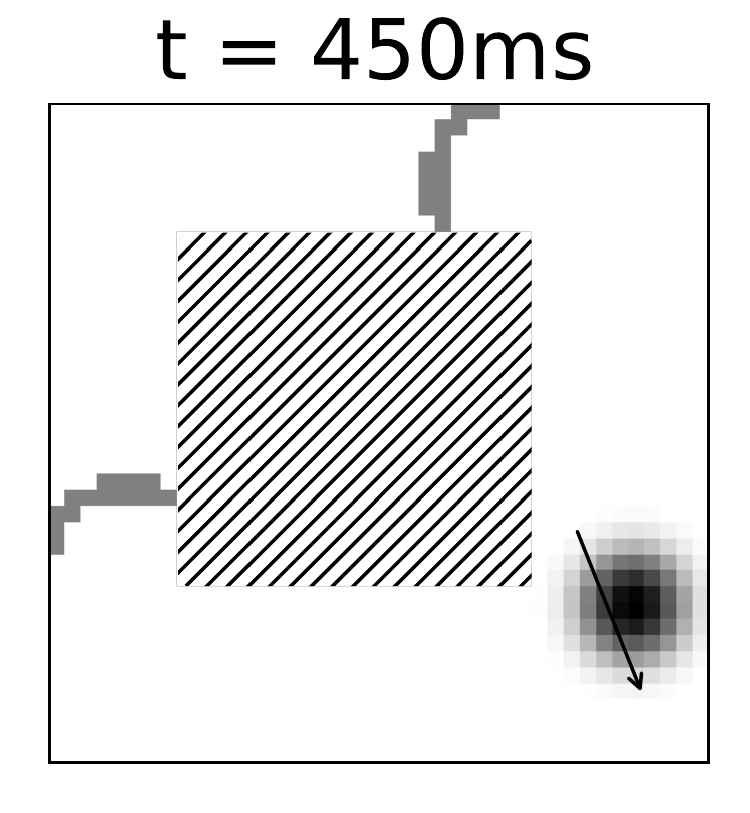}\hfill%
        \includegraphics[width=0.16\textwidth]{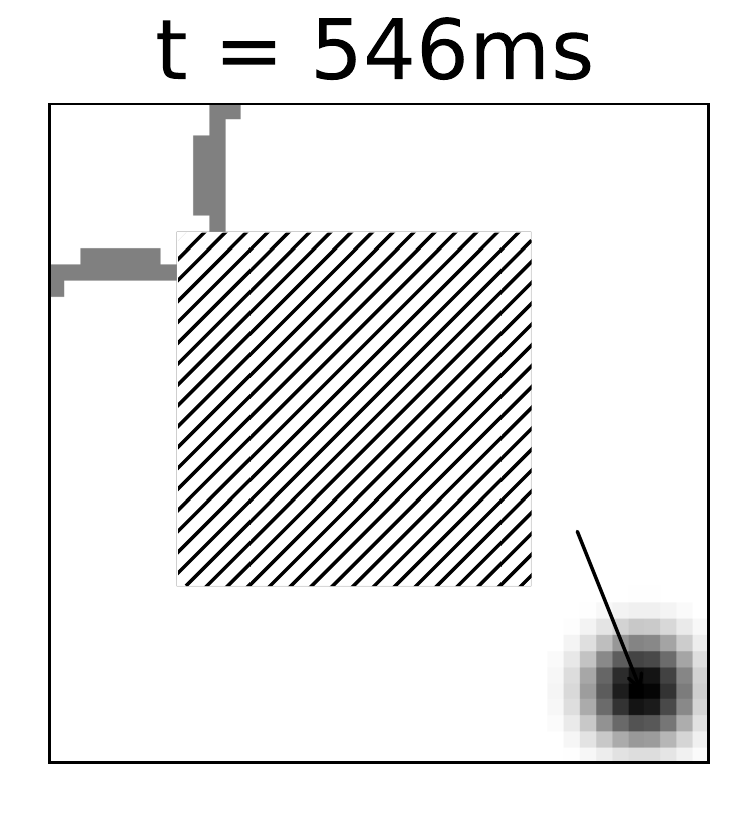}%
        \caption{Block Maze}
        \label{fig:mazes:block}
    \end{subfigure}
    \\[.5em]
    \begin{subfigure}[b]{\textwidth}
        \includegraphics[width=0.16\textwidth]{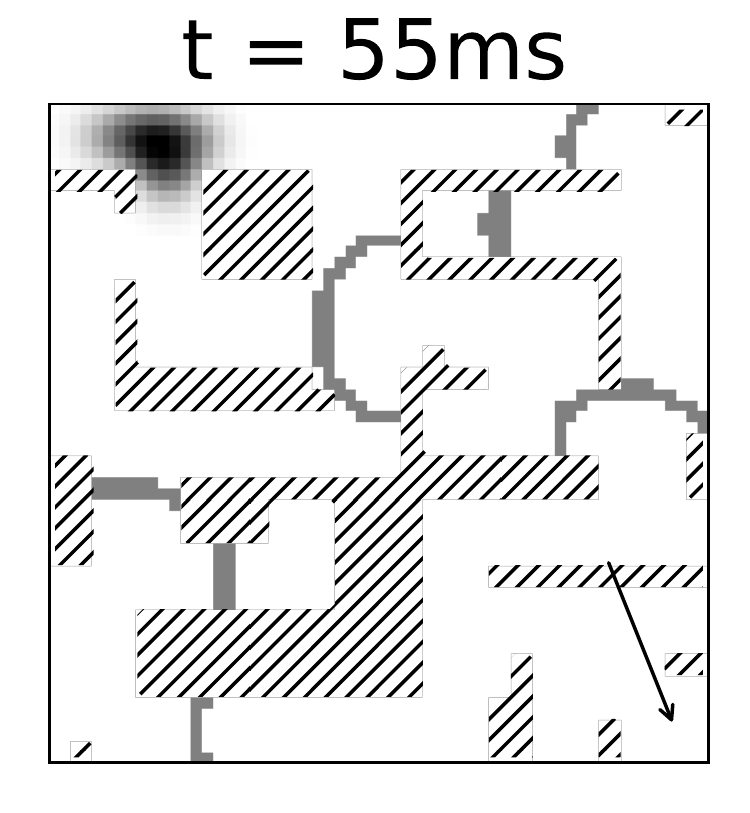}\hfill%
        \includegraphics[width=0.16\textwidth]{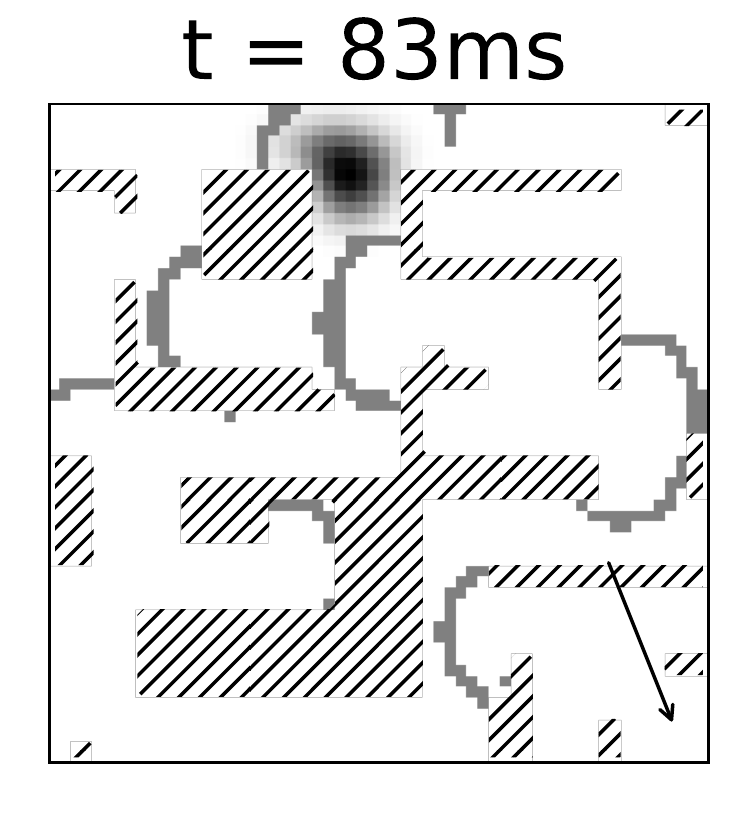}\hfill%
        \includegraphics[width=0.16\textwidth]{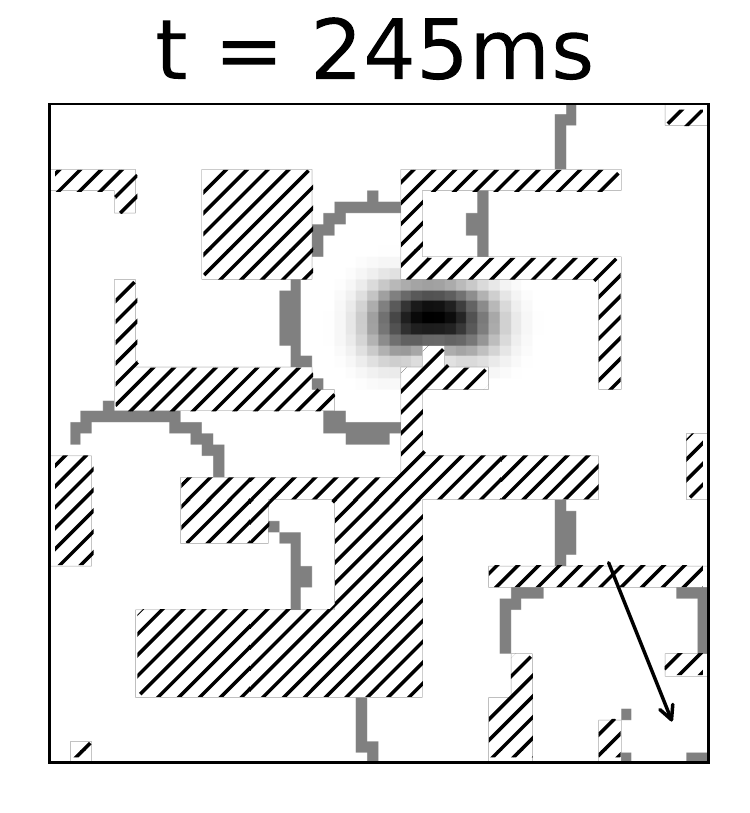}\hfill%
        \includegraphics[width=0.16\textwidth]{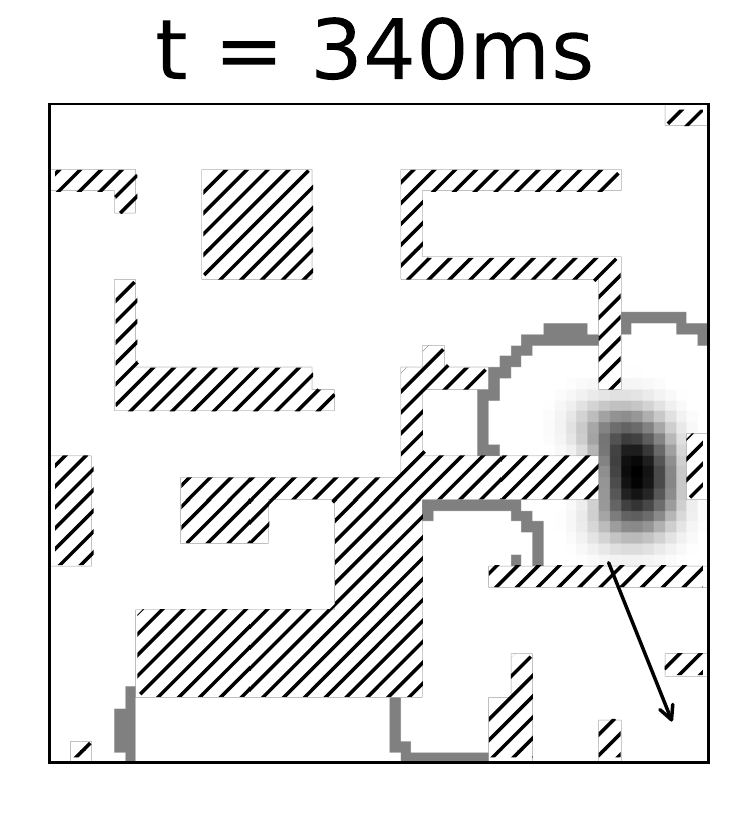}\hfill%
        \includegraphics[width=0.16\textwidth]{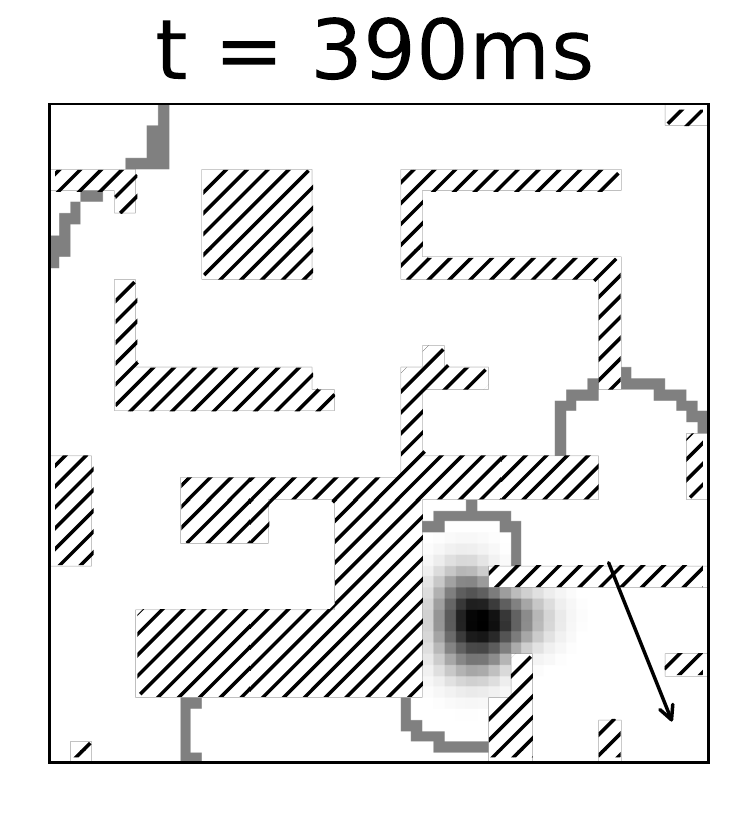}\hfill%
        \includegraphics[width=0.16\textwidth]{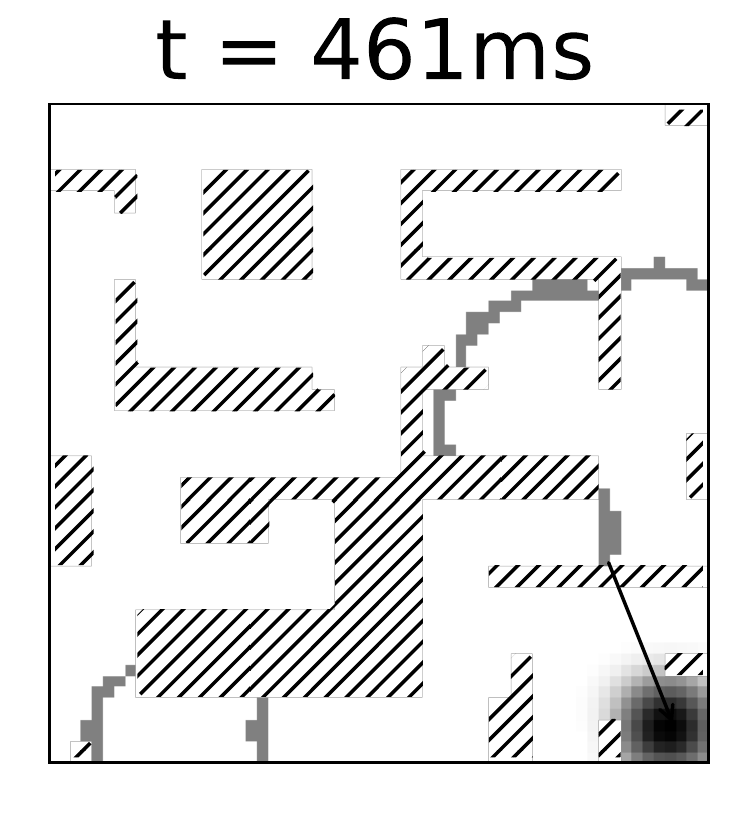}%
        \caption{Complex Maze}
        \label{fig:mazes:complex}
    \end{subfigure}

    \caption{%
            Simulations where specific portions of the neuronal layers were blocked for traversal (hatched regions) show the model's capability of solving complex planning problems.
        }
    \label{fig:mazes}
\end{figure}

As described in \Cref{sec:model:connection}, the manifold of stimuli represented by the neuronal network can be curved, branched, or of different topology, either permanently or temporarily.
The purpose of the model is to allow for a reliable solution to the underlying graph traversal problems independent of potential obstacles in the networks.
For this reason we investigated whether the bump of activation in the continuous attractor layer was able to successfully navigate through the graph from the starting node to the end node in the presence of nodes that could not be traversed.
To test this idea we constructed different \enquote{mazes}, blocking off sections of the graph by zeroing the synaptic connections of the respective neurons in the wave propagation layer and by clamping activation functions of the corresponding neurons in the continuous attractor layer to zero (see \Cref{fig:mazes}).
We found that in all these setups, the algorithm was able to successfully navigate the bump in the continuous attractor layer through the mazes.

\subsection{Relation to Existing Graph Traversal Algorithms} \label{sec:Relation_To_Existing_Algorithms}
To conclude this section, we highlight a few parallels between the presented approach and the classical Breadth-First Search (\BFS) algorithm.

\BFS\, begins at some start node \startnode of the graph and marks this node as \enquote{visited}.
In each step, it then chooses one node which is \enquote{visited} but not \enquote{finished} and checks whether there are still unvisited nodes that have an edge to this node.
If so, the corresponding nodes are also marked as \enquote{visited}, the current node is marked as \enquote{finished} and another iteration of the algorithm is started.
\redact{
    For a more formal treatment of \BFS, we refer to \Cref{app:graph_traversal_algorithms}.
}

The approach presented here is a \emph{parallelized} variant of this algorithm.
Assuming that all neurons always obtain sufficient current to become activated, the propagating wave corresponds to the step of the algorithm in which the neighbors of the currently considered node are investigated.
In contrast to \BFS, the algorithm performs this step for all candidate nodes in a single step.
That is, it considers \emph{all} nodes currently marked as visited, checks the neighbors of all these nodes \emph{at once} and marks them as visited if necessary.
\redact{
    This close connection also allows to derive theoretical performance properties for the algorithm based on the behavior of \BFS.
    As a more in-depth analysis of this connection is not within the scope of this paper, we refrain from going into detail here and refer again to \Cref{app:graph_traversal_algorithms}.
}

Having all ingredients of the proposed conceptual framework in place, the following section reviews some experimental evidence indicating that it could in principle be employed by biological brains.

% =====================================================================
\section{Empirical Evidence} \label{sec:empiric_evidence}
% =====================================================================

\redact{%
In this section we review empirical findings which are relevant for the model.
We dedicate one subsection to each of several key model assumptions on the neural connectivity and dynamics.
}
% =====================================================================
\subsection{Cognitive Maps} \label{sec:cognitive_maps}
% =====================================================================

The concept of \enquote{cognitive maps} was first proposed by Edward Tolman \cite{Tolman1930,Tolman1948} who conducted experiments to understand how rats were able to navigate mazes to seek rewards.
\redact{
    He noticed that these animals showed remarkably flexible behaviour when confronted with novel versions of their maze environments, such as finding previously unexplored shortcuts or finding new routes when obstructions made old ones untraversable.
    Tolman theorized that this behaviour was made possible by the rats having an internal model (or map) of the mazes which they used to navigate and which they updated when new information about the maze was presented.
}

A body of evidence suggests that neural structures in the hippocampus and enthorinal cortex potentially support cognitive maps used for spatial navigation \cite{Okeefe1971, Okeefe1978, Bush2015}.
Within these networks, specific kinds of neurons are thought to be responsible for the representation of particular aspects of cognitive maps.
Some examples are place cells \cite{Okeefe1971, Okeefe1978} which code for the current location of a subject in space, grid cells which contribute to the problem of locating the subject in that space \cite{Hafting2005} as well as supporting the stabilisation of the attractor dynamics of the place cell network \cite{Guanella2007}, head-direction cells \cite{Taube1990} which code for the direction in which the subject's head is currently facing, and reward cells \cite{Gauthier2018} which code for the location of a reward in the same environment.

The brain regions supporting spatially aligned cognitive maps might also be utilized in the representation of cognitive maps in non-spatial domains:
In \cite{Constantinescu2016}, fMRI recordings taken from participants while they performed a navigation task in a non-spatial domain showed that similar regions of the brain were active for this task as for the task outlined in \cite{Doeller2010} where participants navigated a virtual space using a VR apparatus.
\redact{Not only were the same spatial-task aligned regions active for this non-spatial-domain navigation task but the firing patterns of the neurons recorded in the former displayed the same hexagonal firing patterns that are characteristic of enthorinal grid-cells.}
Further, according to \cite{Cameron2001}, activation of neurons in the hippocampus (one of the principal sites for place cells) is indicative of how well participants were able to perform in a task related to pairing words.
Supporting this observation with respect to the role played by these brain regions in the operation of abstract cognitive maps, \cite{Alvarez2002} found that lesions to the hippocampus significantly impaired performance on a task of associating pairs of odors by how similar they smelled.
Finally, complementing these findings, rat studies have shown that hippocampal cells can code for components in navigation tasks in auditory \cite{Aronov2017, Sakurai2002}, olfactory \cite{Eichenbaum1987}, and visual \cite{Fried1997} task spaces.

\redact{
    Taken as a whole, the above body of research provides good evidence for the following ideas:
    Firstly, that cognitive maps exist in humans.
    Secondly, that these maps can and are used for solving problems in a general class of task spaces.
    Thirdly, that hippocampal and enthorinal cells likely play a key role in the construction and operation of these maps.
}

% =====================================================================
\subsection{Feed-Forward and Recurrent Connections} \label{sec:correlation_graph}
% =====================================================================

As described in \Cref{sec:model:manifold}, the proposed model is built around a particular \emph{theme of connectivity}:
Each neuron represents a certain pattern in sensory perception mediated via feed-forward connections.
\redact{Such a pattern could be, for example, all the percepts associated with a particular position in a maze, a certain body posture, or some letter in the field of vision, \cf \Cref{fig:examples}.}
In addition, recurrent connections between two neurons strengthen whenever they are activated simultaneously. In the following, we give an overview of some relevant experimental observations which are consistent with this mode of connectivity.

The most prominent example of neurons which are often interpreted as pattern detectors are the cells in primary visual cortex.
These neurons fire when a certain pattern \redact{(like a small edge of bright/dark contrast)} is perceived at a particular position and orientation in the visual field. On the one hand, these neurons receive their feed-forward input from the lateral geniculate nucleus.
On the other hand, they are connected to each other through a tight network of recurrent connections. Several studies (see \eg \cite{ko_emergence_2013, iacaruso_synaptic_2017, ko_functional_2011}) have shown that two such cells are preferentially connected when their receptive fields are co-oriented and co-axially aligned.
Due to the statistical properties of natural images, where elongated edges appear frequently, such two cells can also be expected to be positively correlated in their firing due to feed-forward activation.

\redact{
    Similar statements are valid for auditory cortex: Neurons in primary auditory cortex receive feed-forward input from thalamocortical connections as well as intracortical signals via recurrent connections. The feed-forward input is tonotopically organized and A1 neurons typically respond to one or several characteristic frequencies. There is evidence for a cross-frequency integration via intracortical input:
    For example, neurons in A1 show subthreshold responses to frequency ranges broader than can be accounted for by their thalamic inputs \cite{kaur_intracortical_2004} while the latency of their response is shortest at their characteristic frequency \cite{kaur_spectral_2005}. Additional supporting facts are reviewed in the introduction of \cite{kratz_spatial_2015}. By analogy from the visual cortex, one might expect that intra-cortical connections are strongest between neurons if their characteristic frequencies differ by a harmonic interval, \eg by a full octave, since such intervals are most highly correlated in the frequency spectra of natural sounds \cite{Abdallah:2006a, Abdallah:2006b}. While we are not aware of any study examining this particular conjecture, there is a lot of evidence that harmonics play a major role in the organization of the auditory cortex in general \cite{wang_harmonic_2013}.
}

The somatosensory cortex is another brain region where several empirical findings are in line with the postulated theme of 
\redact[connectivity. Experiments]{connectivity. Area 3b in the somatosensory cortex contains neurons which respond to tactile stimuli.
Their receptive fields are not dissimilar to those of cells in V1. Experiments} on non-human primates suggest that \enquote{3b neurons act as local spatiotemporal filters that are maximally excited by the presence of particular stimulus features} \cite{dicarlo_structure_1998}.

Regarding the recurrent connections in somatosensory cortex, some empirical support stems from the well-studied rodent barrel cortex.
Here, the animal's facial whiskers are represented somatotopically by the columns of primary somatosensory cortex.
Neighboring columns of the barrel cortex are connected via a dense network of recurrent connections.
Sensory deprivation studies indicate that the formation of these connections depends on the feed-forward activation of the respective columns:
If the whiskers corresponding to one of the columns are trimmed during early post-natal development, the density of recurrent connections with this column is reduced \cite{wallace_plasticity_2008, broser_critical_2008}.
Conversely, synchronous co-activation over the course of a few hours can lead to increased functional connectivity in the primary somatosensory cortex \cite{vidyasagar_re-wiring_2014}.

The primary somatosensory cortex also receives proprioceptive signals from the body which represent individual joint angles.
Taken as a whole, these signals characterize the current posture of the animal and there is an obvious analogy to the arm example, \cf \Cref{fig:examples:robot}.
We are not aware of any experimental results regarding the recurrent connections between proprioception detectors, but it seems reasonable to expect that the results about processing of tactile input in the somatosensory cortex can be extrapolated to the case of proprioception.
This would imply that a recurrent network structure roughly similar to \Cref{fig:examples:robot} should emerge and thus support the model for controlling the arm.

Area 3a of the somatosensory cortex, whose neurons exhibit primarily proprioceptive responses, is also densely connected to the primary motor cortex.
It contains many corticomotoneuronal cells which drive motoneurons of the hand in the spinal cord \cite{delhaye_neural_2018}.
This tight integration between sensory processing and motor control might be a hint that the hypothetical string-of-a-puppet muscle control mechanism from \Cref{sec:model:connection} is not too far from reality.

In summary, evidence from primary sensory cortical areas seems to suggest a common cortical theme of connectivity in which neurons are tuned to specific patterns in their feed-forward input from other brain regions, while being connected intracortically based on statistical correlations between these patterns.

% =====================================================================
\subsection{Wave Phenomena in Neural Tissue} \label{sec:waves}
% =====================================================================

\redact{
In the model we present, the target state of a cognitive planning task is encoded by localized activation within the cognitive map.
Starting from there, neural activation travels through the recurrent network in what resembles expanding wave fronts.
}

There is a large amount of empirical evidence for different types of wave-like phenomena in neural tissue.
We summarize some of the experimental findings, focusing on fast waves (a few tens of \si{\centi\meter\per\second}). %which travel horizontally through cortical areas over a distance of several millimeters.
These waves are suspected to have some unknown computational purpose in the brain \cite{muller_cortical_2018} and they seem to bear the most resemblance with the waves postulated in the model.

Using multielectrode local field potential recordings, voltage-sensitive dye, and multiunit measurements, traveling cortical waves have been observed in several brain areas, including motor cortex, visual cortex, and non-visual sensory cortices of different species. There is evidence for wave-like propagation of activity both in sub-threshold potentials and in the spatiotemporal firing patterns of spiking neurons \cite{Sato:2012}.

In the motor cortex of wake, behaving monkeys, Rubino \etal \cite{rubino_propagating_2006} observed wave-like propagation of local field potentials.
They found correlations between some properties of these wave patterns and the location of the visual target to be reached in the motor task.
On the level of individual neurons, Takahasi \etal found a \enquote{spatiotemporal spike patterning that closely matches propagating wave activity as measured by LFPs in terms of both its spatial anisotropy and its transmission velocity} \cite{takahashi_large-scale_2015}.

In the visual cortex, a localized visual stimulus elicits traveling waves which traverse the field of vision. For example, Muller \etal have observed such waves rather directly in single-trial voltage-sensitive dye imaging data measured from awake, behaving monkeys \cite{muller_stimulus-evoked_2014}.

\redact{
    The detailed propagation mechanisms which lead to fast travelling waves in cortical tissue are still under discussion.
    The prevalent view seems to be that waves are actually propagated through the circuitry of the respective cortical area rather than, being the result of spatiotemporally organized activation that stems from some other brain region.
    Two competing mechanisms for waves \cite{Sato:2012} are: (1) strictly localized generation of activity followed by monosynaptic propagation through long-range horizontal connections of the superficial cortical layers or (2) a \enquote{chain reaction} of firing neurons leading to a self-sustaining spread of activity through the deeper cortical layers.
    While possibly both mechanisms play a role in the brain, only the second one is incorporated in the model.
}

% =====================================================================
\subsection{Spatial Navigation Using Place Cells} \label{sec:spatial_navigation}
% =====================================================================

Finding a short path through a maze-like environment, \cf \Cref{fig:examples:maze}, is one of the planning problems the model is capable of solving.
In this case, each neuron of the continuous attractor layer represents a \enquote{place cell} which encodes a particular location in the maze.

Place cells were discovered by John O'Keefe and Jonathan Dostrovsky in 1971 in the hippocampus of rats \cite{okeefe_hippocampus_1971}.
They are pyramidal cells that are active when an animal is located in a certain area (\enquote{place field}), of the environment.
Place cells are thought to use a mixture of external sensory information and stabilizing internal dynamics to organize their activity:
On the one hand, they integrate external environmental cues from different sensory modalities to anchor their activity to the real world.
This is evidenced by the fact that their activity is affected by changes in the environment and that it is stable under a removal of a subset of cues \cite{barry_boundary_2006, jeffery_place_2011}.
On the other hand, firing patterns are then stabilized and maintained by internal network dynamics as cells remain active under conditions of total sensory deprivation \cite{quirk_firing_1990}. Collectively, the place cells are thought to form a cognitive map of the animal's environment.

\redact{
In theoretical or computational studies, continuous attractor models are often used to describe place cell dynamics. Just as we do in the present article, it is typically assumed that each place cell responds, on the one hand, to location-specific patterns of sensory cues and, on the other hand, to stimulation via recurrent connections from cells with overlapping place fields.
}

% =====================================================================
\subsection{Targeted Motion Caused by Localized Neuron Stimulation} \label{sec:stimulation}
% =====================================================================

\redact{
In our model, the process of motion planning is triggered by stimulating the neurons which represent the body's to-be position, \cf \Cref{fig:examples:robot}.
In the present section, we review some experimental results that support the biological plausibility of this assumption.
}

In 2002, Graziano \etal reported results from electrical microstimulation experiments in the primary motor and premotor cortex of monkeys \cite{Graziano:2002}.
Stimulation of different sites in the cortical tissue for a duration of \SI{500}{\milli\second} resulted in complex body motions involving many individual muscle commands.
The stimulation of one particular site typically led to smooth movements with a certain end state, independent of the initial posture of the monkey, while stimulating a different location in the cortical tissue led to a different end state.
\redact{
In particular, Graziano \etal noted that stimulation at a fixed site can have the different effects in terms of low-level muscle commands:
For example, a monkey's arm might either stretch or flex to reach a partially flexed position, depending on its initial condition.
}
In terms of the model presented here, this would be explained by two wave fronts propagating in opposite directions away from the to-be location, only one of which hits the localized peak of activity encoding the as-is location and pulls it closer to the to-be state.
Graziano \etal also reported that the motions stopped as soon as the electrical stimulus was turned off. This is fully consistent with our model, where stopping the to-be activation means that no more wave fronts are created and thus the as-is peak of activity remains where it is.

After this original discovery by Graziano \etal in 2002, several additional studies have confirmed and extended their results, see \cite{graziano_ethological_2016} for an overview.
\redact{
Similar effects of motor cortex stimulation have been observed in a variety of different primate and rodent species.
The results also hold true for different types of neural stimulation: electrical, chemical and optogenetic.
}
The neural structures which cause the bodily motions towards a specific target state have been named \emph{ethological maps} or \emph{action maps} \cite{graziano_ethological_2016}.

Furthermore, several studies suggest that such action maps are shaped by experience:
Restricting limb movements for thirty days in a rat can cause the action map to deteriorate. %, as indicated by a reduced number of sites where complex movements of the respective limb can be evoked.
A recovery of the map is observed during the weeks after freeing the restrained limb \cite{budri_sensorimotor_2014}. Conversely, a reversible local deactivation of neural activity in the action map can temporarily disable a grasping action in rats \cite{brown_motor_2014}. A permanent lesion in the cortical tissue can disable an action permanently. The animal can re-learn the action, though, and the cortical tissue reorganizes to represent the newly re-learnt action at a different site \cite{ramanathan_form_2006}. These observed plasticity phenomena are fully in line with our model which emphasises a self-organized formation of the cognitive map via Hebbian processes both for the feature learning and for the construction of the recurrent connections.

% =====================================================================
\subsection{Participation of the Primary Sensory Cortex in Non-Sensory Tasks} \label{sec:planning_tasks}
% =====================================================================

For the first two examples in \Cref{fig:examples}, the association with a planning task is obvious.
Our third example, the geometric transformations of the letter \enquote{A}, may appear a bit more surprising, though:
After all, the neural structures in visual sensory cortex would then be involved in \enquote{planning tasks}.
The tissue of at least V1 fits the previously explained theme of connectivity, but it is often thought of as a pure perception mechanism which aggregates optical features in the field of vision and thus performs some kind of preprocessing for the higher cortical areas.

However, there is evidence that the visual sensory cortex plays a much more active role in cognition than pure feature detection on the incoming stream of visual sensory information.
In particular, the visual cortex is active in visual imagery, that is, when a subject with closed eyes mentally imagines a visual stimulus \cite{pearson_human_2019}.
\redact{Experiments suggest that mental imagery leads to activation patterns in the early visual cortex which are composed of the same visual features as during actual sensory perception:
Using multi-voxel pattern classification on fMRI measurements of the visual cortex, it is possible to train machine learning models which can accurately decode cortical activation and determine which image in the field of vision has caused the neural response.
The same models, trained only on perceptual images, have been used successfully to decode cortical activation caused by purely mental images \cite{naselaris_voxel-wise_2015}.}

Based on such findings, it has been suggested that \enquote{the visual cortex is something akin to a \enquote{representational blackboard} that can form representations from either the bottom-up or top-down inputs} \cite{pearson_human_2019}.
In our model, we take this line of thinking one step further and speculate that the early visual cortex does not only represent visual features, but that it also encodes possible transformations like rotation, scaling or translation via its recurrent connections.
In this view, the \enquote{blackboard} becomes more of a \enquote{magnetic board} on which mental images can be placed and shifted around according to rules which have been learned by experience.

Of course, despite the over-simplifying \Cref{fig:examples:A}, we do not intend to imply that there were any neurons in the visual cortex with a complex pattern like the whole letter \enquote{A} as a receptive field.
In reality, we would expect the letter to be represented in early visual cortex as a spatio-temporal multi-neuron activity pattern.
The current version of our model, on the other hand, allows for single-neuron encoding only and thus reserves one neuron for each possible position of the letter.
We will discuss this and other limitations of the proposed model in \Cref{sec:discussion}.

% =====================================================================
\subsection{Temporal Dynamics} \label{sec:timing}
% =====================================================================

The concept presented in this article implies predictions about the temporal dynamics of cognitive planning processes which can be compared to experiments: The bump of activity only starts moving when the first wave front arrives. Assuming that every wave front has a similar effect on the bump, its speed of movement should be proportional to the frequency with which waves are emitted. Thus both the time until movement onset and the duration of the whole planning process should be proportional to the length of the traversed path in the cortical map. Increased frequency of wave emission should accelerate the process.

One supporting piece of evidence is provided by mental imagery: Experiments in the 1970s \cite{shepard_mental_1971,cooper_chronometric_1973} have triggered a series of studies on mental rotation tasks, where the time to compare a rotated object with a template has often been found to increase proportionally with the angle of rotation required to align the two objects.

In the case of bodily motions, the total time to complete the cognitive task is not a well suited measure since it strongly depends on mechanical properties of the limbs. Yet for electrical stimulation of the motor cortex (\cf \Cref{sec:stimulation}) Graziano \etal report that the speed of evoked arm movements increases with stimulation frequency \cite{graziano_arm_2005}. Assuming that this frequency determines the rate at which the hypothetical waves of activation are emitted, this is consistent with our model. 

In addition, our model makes the specific prediction that the latency between stimulation and the onset of muscle activation should increase with the distance between initial and target posture. We are not aware of any studies having examined this particular relationship yet.%, but in general the latency between motor cortex stimulation and an EMG response in monkeys has been found to be roughly 7ms, only a fraction of which could be attributed to a hypothetical wave propagation in cortex. This determines an upper bound for the time of wave propagation at a few ms.

% =====================================================================
\section{Discussion}\label{sec:discussion}
% =====================================================================

The model proposed here is, to the best of our knowledge, the first model that allows for solving graph problems in a biological plausible way such that the solution (\ite the specific path) can be calculated directly on the neuronal network as the only computational substrate.

Similar approaches and models have been investigated earlier, especially in the field of neuromorphic computing.
For example, in \cite{muller_hippocampus_1996,Aimone:2019,Aimone:2021,Hamilton:2019,Kay:2020} graphs are modeled using neurons and synapses, and computations are performed by exciting specific neurons which induces propagation of current in the graph and observing the spiking behavior.
Although some models are more general than the one presented here and allow for solving more complex problems like dynamic programs \cite{Aimone:2019}, enumeration problems \cite{Hamilton:2019} or the longest shortest path problem \cite{Kay:2020}, we are not aware of any model explicitly discussing the biological plausibility.
In fact, most of these approaches are far from being biologically plausible as they \eg require additional artificial memory \cite{Aimone:2019} or a preprocessing step that changes the graph depending on the input data \cite{Kay:2020}.
Also, the model of Muller \etal \cite{muller_hippocampus_1996} as well as the very recent model of Aimone \etal \cite{Aimone:2021} which are biologically more plausible do not discuss how a specific path can then be computed in the graph, even if the length of a path can be calculated \cite{Aimone:2021}.

\redact{
    Our model has not been created with the intent to explain empirical findings from one particular brain region, mental task or experimental technique in full detail. Rather, we sought to explore ways in which a generic algorithmic framework might solve seemingly very different problems based on more or less the same neural substrate. Working on a relatively high level of abstraction and ignoring most of the domain-specific aspects may not only help our understanding of computational principles employed by the brain but also pave the road to the development of new useful algorithms in artificial intelligence. Nevertheless, it is important to note that many features of our model are in line with experimental results from various areas of brain science and we review those findings in \Cref{sec:empiric_evidence}.
}

In the following we discuss limitations of the presented model and potential avenues for further research.

\subsection{Single-Neuron vs. Multi-Neuron Encoding}
In our model, each point on a cortical map is represented by a single neuron and a distance on the map is directly encoded in a synaptic strength between two neurons. 
The graph of synaptic connections an therefore be considered as a coarse-grained version of the underlying manifold of stimuli. %, making the model intuitive and relatively convenient to implement. 
Yet such a single-neuron representation is possible only for manifolds of a very low dimension, since the number of points necessary to represent the manifold grows exponentially with each additional dimension. For tasks like bodily movement, where dozens of joints need to be coordinated, the number of neurons required to represent every possible posture in a single-neuron encoding is prohibitive. Therefore, it is desirable to encode manifolds of stimuli in a more economical way -- for example, by representing each point of the manifold by a certain set of neurons. It is an open question how distance relationships between such groups of neurons could be encoded and whether the dynamics from our model could be replicated in such a scenario.

\redact{
    \subsection{Wave Propagation and Continuous Attractor Layers}\label{sec:discussion:merging_layers}
    
    Certain design choices made in the numerical implementation should be discussed regarding their biological plausibility and possible alternative mechanisms.
    
    If the wave propagation layer and the continuous attractor layer were to form organically as two separate sub-circuits in a real biological system, each of their neurons would need to act as a feature detector, since otherwise it is not clear how the right structure of recurrent connections could develop. Then each feature will be represented by two detectors -- one in each layer -- and there must be some unknown mechanism which establishes a link between every pair of corresponding feature detectors. 
    
    Moreover, the split of the model dynamics into two layers leads to a somewhat artificial implementation of the interaction between them:
    As described in \Cref{sec:model:implementation}, we compute the direction into which the activation bump in the continuous attractor layer is shifted whenever a wave front arrives at the corresponding location in the wave propagation layer.
    The details of this mechanism do not appear to be biologically plausible and we would rather expect that the bump is moved only by the aggregated effects of local interactions between synaptically connected neurons. This interaction could be mediated by the connections between corresponding feature detectors which we postulated above.
    
    Alternatively, an elegant and biologically plausible model could be obtained by merging the wave propagation and continuous attractor dynamics into a single layer of neurons.
    In such a single-layer model, the whole network can form in a self-organized way: First, the feed-forward connections are generated via a process of competitive Hebbian learning, leading to a network of individual feature detectors.
    In a second step, these detectors establish recurrent connections among each other, again driven by Hebbian learning, to create the graph structure required in our model.

    In the single-layer scenario, the model must allow for continuous attractor dynamics and wave-like expansion of activity simultaneously.
    We speculate that this is possible in principle, for example by imposing a time delay on the effect of inhibition -- which appears biologically plausible considering that it is mediated in an extra step via inhibitory interneurons.
    The time delay of inhibition has only minimal effect on the quasi-static peak of activity and thus conserves the landscape of continuous attractors from the two-layer scenario.
    On the other hand, the time delay allows activation patterns with strong temporal fluctuations to emit waves of activity before inhibition has any effect.

    The interaction between the waves and the localized peak of activity could potentially shift the peak in the direction of the incoming waves without the need to impose any artificial assumptions to the model:
    The incoming waves are annihilated by the peak's \enquote{trench of inhibition}, but they also increase the level of activation of the neurons on the side of the bump which is hit by the wave.
    Due to the attractor dynamics of the network the bump recovers from this deformation, but in the process it changes its location slightly towards the direction of the incoming wave.

    Realizing the effects described above will require a very careful numerical set-up of the model and tuning of its parameters.
    We consider this an interesting direction for future research since the potential outcome is a rather elegant model with a high degree of biological plausibility.
}

\subsection{Embedding into a Bigger Picture}

While the model focuses on the solution of graph traversal problems, it appears desirable to embed it into a broader context of sensory perception, decision making, and motion control in the brain.
One particular question is how the hypothetical \enquote{puppet string mechanism} -- which we postulated to connect proprioception and motion control -- could be implemented in a neural substrate.
Similarly, if our model provides an appropriate description of place cells and their role in navigation, the question arises how a shift in place cell activity is translated into appropriate muscle commands to propel the animal in the corresponding direction. 

It is intriguing to speculate about a deeper connection between our model and object recognition:
On the same neural substrate, our hypothetical waves might travel through a space of possible transformations, starting from a perceived stimulus and \enquote{searching} for a previously learned representative of the same class of objects.
This could explain why recognition of rotated objects is much faster than the corresponding mental rotation task \cite{corballis_decisions_1978}:
The former would require only one wave to travel through the cognitive map, while the later would require many waves to move the bump of activity.

\redact{
    And finally, an open question is the connection between the model and the hypothetical executive brain functions which are assumed to define the target state for the graph traversal problem and activate the corresponding neurons. 
}
% =====================================================================
\section{Conclusion}\label{sec:conclusion}
% =====================================================================

We have shown that a wide range of cognitive tasks, especially those that involve planning, can be represented as graph problems. 
To this end, we have detailed one possible role for the recurrent connections that exist throughout the brain as computational substrate for solving graph traversal problems.
We showed in which way such problems can be modeled as finding a short path from a start node to some target node in a graph that maps to a manifold representing a relevant task space.
Our review of empirical evidence indicates that a theme of connectivity can be observed in the neural structure throughout (at least) the neocortex which is well suited to realize the proposed model.

\redact{
We constructed a two-layer neural network consisting of a layer of neurons that implemented a continuous attractor sheet modelled after neurons found in the human hippocampus and enthorinal cortex.
This sheet of neurons enacts a "bump" of activity centered on the neuron representing the start node \startnode in the graph.
As a second step we implemented a sheet of spiking neurons that generated a wave of activation across the same sheet starting from a individual neuron that represented the target node \targetnode in the graph.
Finally, we implemented an interaction algorithm which caused the bump of activation in the continuous attractor layer to move in the direction of the wave front as it reached the region of activation in the continuous attractor layer.
We found that this model was successfully able to move the activation bump in the continuous attractor layer through the sheet of neurons to the location that mapped to the target node \targetnode in the graph, thus solving the graph traversal problem. We found further that the model was robust to large changes in the graph structure.
Specifically we showed that if large portions of the graph are made inaccessible and the relevant neurons in the model were zeroed out that the model is still able to guide the activation bump from the start node \startnode to the target node \targetnode successfully.

Despite its relatively small scale we believe that models such as ours may provide a starting point in understanding how brains are able to exhibit flexible behaviour with respect to different kinds of cognitive tasks.
Apparently, the stereotypical theme of connectivity encountered across the neocortex allows the brain to create a model of its environment based on sensory perception.
Once established, the neural structure can be used as a \enquote{planning board} to support different cognitive tasks.

Next to a deeper understanding of the brain, we believe that models like ours can be an inspiration for new algorithms of artificial intelligence (AI). Artificial neural networks used in technical applications today are typically input-driven, they rely on feed-forward processing of information through several layers of neurons, and they are trained via supervised learning.
The brain, however, continuously integrates sensory input into its own dynamics, its connectivity structure is mostly recurrent, and learning happens to a large extent in an unsupervised way.
In all three of these fundamental differences, our model is aligned more closely to the properties of the brain than those of most other AI algorithms.
At the same time, it shows how relevant computational problems can be solved with a very generic approach that relies heavily on self-organization. Potential applications include motion control for robots, especially in scenarios which require a high degree of flexibility and continuous adaptation to changing circumstances. As an additional interesting feature, since the model is based on artificial spiking neurons, it can potentially be implemented very efficiently in neuromorphic computer hardware.
}

\section*{Acknowledgements}
The authors gratefully acknowledge funding from the European Research Council (ERC) under the European Union’s Horizon 2020 research and innovation program (Grant agreement 677270) which allowed H.\,P. to undertake the work described in the above.
We also thank Raul Mure\textcommabelow{s}an and Robert Klassert for providing very valuable comments on early drafts of the paper.

\section*{Methods and Experiments}\label{sec:methods}

\redact[
    The model described in \Cref{sec:model} treats the recurrent neural network as a discretized approximation to the manifold of stimuli.
    Thus, the problem of finding a short path through that manifold  translates into a graph traversal problem in the corresponding graph of synaptic connections.
    In the following, the starting and target position of the planning process are denoted by \startnode and \targetnode, respectively.
]{
    \subsection*{Connection to Mathematical Graph Traversal Problems}
    As the model described in \Cref{sec:model} uses a neural network of neurons to solve planning problems in the cognitive map, it is natural to interpret this network as a graph consisting of nodes representing the neurons and edges representing their synaptic connections.
    Thus, the planning problem in the network translates into a graph traversal problem in the corresponding graph.
    In the following, we hence introduce some basic terminology used in the field of graph theory.

    We refrain from giving too many details and references, as most of the standard formalism can be found in classical books on mathematical optimization.
    In particular, we refer to \cite{Korte:2018,Schijver:2003} for references, details, proofs and further discussions.

    A \emph{graph} $G$ is a pair $G=(\V,\E)$, where $\V$ is a finite set of \emph{nodes} and $\E$ is the set of \emph{edges}, where each edge is set of two nodes.
    For two nodes $\startnode,\targetnode\in\V$, an \emph{$\startnode$-$\targetnode$-path} is a path of nodes starting at $\startnode$  and ending at the target node $\targetnode$ such that any two consecutive nodes along the path are connected by an edge.
    %Formally, it is a finite sequence of nodes $(\startnode=v_1,v_2,\dots,v_k=\targetnode)$ such that $\{v_i,v_{i+1}\}\in\E$ for all $i\in\{1,\dots,k-1\}$.
    A node $\targetnode\in\V$ is \emph{reachable} from another node $\startnode\in V$ if there exists an $\startnode$-$\targetnode$-path in $G$.
    An example of a graph with 15 nodes and different paths is given in \Cref{fig:GraphExample}.
}

\redact{
    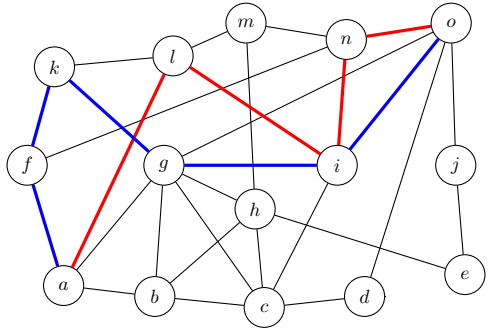
\begin{figure}[H]
        \centering
        \begin{tikzpicture}
        \tikzset{every node/.style={draw, circle, inner sep=2pt, scale=0.66, minimum size=0.8 cm}}
        \def\x{1.2}
        \def\y{1.45}
        \node (v1) at (-0.1*\x,-0.1*\y) {$a$};
        \node (v2) at (0.9*\x,-0.2*\y) {$b$};
        \node (v3) at (2.1*\x,-0.3*\y) {$c$};
        \node (v4) at (3.2*\x,-0.2*\y) {$d$};
        \node (v5) at (4.3*\x,0) {$e$};
        \node (v6) at (-0.5*\x,\y) {$f$};
        \node (v7) at (\x,\y) {$g$};
        \node (v8) at (2*\x,0.6*\y) {$h$};
        \node (v9) at (2.9*\x,\y) {$i$};
        \node (v10) at (4.2*\x,\y) {$j$};
        \node (v11) at (-0.2*\x,1.9*\y) {$k$};
        \node (v12) at (1.1*\x,2*\y) {$l$};
        \node (v13) at (1.9*\x,2.3*\y) {$m$};
        \node (v14) at (3*\x,2.15*\y) {$n$};
        \node (v15) at (4.15*\x,2.3*\y) {$o$};
    
        \foreach \i/\j in {1/2, 1/6, 1/7, 1/12, 2/7,2/8,2/3,3/7,3/8,3/9,3/4,4/4,4/15,5/10,5/8,6/11,6/14, 7/11,7/8,7/9,7/15,8/13,9/12,9/14,9/15,10/15,11/12,12/13,13/14,14/15}{
            \draw (v\i)--(v\j);
        }
    
        \draw[red, very thick] (v1)--(v12)--(v9)--(v14)--(v15);
        \draw[blue, very thick] (v1)--(v6)--(v11)--(v7)--(v9)--(v15);
    
        \end{tikzpicture}
        \caption{An example of a graph $G$ with 15 nodes.
        The red resp. blue edges show two $a$-$o$-paths in the graph.} \label{fig:GraphExample}
    \end{figure}
}

\redact{
    In the following, we let $G=(\V,\E)$ be a fixed graph.
    For simplicity, we assume that every node is reachable from every node.

    We are interested in finding a path between two given nodes $\startnode,\targetnode\in\V$ in $G$.
    The idea is that the node $\targetnode$ represents the neuron encoding the to-be state and $\startnode$ represents the neuron encoding the as-is state of the underlying planning problem.
    To formalize this problem, we denote by \texttt{Path}$(\startnode,\targetnode)$ the problem of finding a path from $\startnode$ to $\targetnode$ for given nodes $\startnode,\targetnode$.

    Even though this problem technically only asks for finding \emph{some} path from $\startnode$ to $\targetnode$, shorter paths that use as few connections as possible are superior to longer paths using more connections. The reason is that the fewer connections a path has, the fewer intermediate states are traversed in the planning problem.
    When considering the previous example of grabbing a cup of coffee, a possible solution could be to move the arm around the head before performing actually reaching towards the coffee cup.
    This is not the movement that would be performed in actual behavior.
    However, we are similarly not obliged to find the shortest possible path.
    Considering the previous example again, a shortest path would reflect a movement with as few intermediate positions as possible.
    This might correspond to stretching the arm in such a way that the cup can barely be reached and might yield an unrealistic behavior.
    Thus, in summary, our goal is to find reasonably \emph{short} paths that do not necessarily need to be \emph{shortest} paths.

    The \texttt{Path}$(\startnode, \targetnode)$ problem is a well-investigated problem in computer science and mathematics.
    With \BFS and \emph{DFS}, two standard path finding algorithms from computer science are described in \Cref{app:graph_traversal_algorithms}.
    There, we also argue why these approaches cannot directly be applied to our scenario due to the fact that the graphs we consider represent neural networks in which algorithms have to be performed in a biological plausible way.
}

\redact{
% =====================================================================
\subsection*{Mathematical Background and Solving the \texttt{Path} Problem in Typical Graphs} \label{app:graph_traversal_algorithms}
% =====================================================================

In the following, we consider how the \texttt{Path}$(\startnode,\targetnode)$ problem can be solved in general graphs that do not represent neural networks.
We later discuss what problems occur when trying to adapt these algorithms to such graphs when using the neurons as computational substrate.
In all of the following, we omit technical details and proofs and instead refer to \cite{Schijver:2003,Korte:2018} again.

Consider some fixed graph $G=(\V,\E)$ and two nodes $s,t\in\V$.
For simplicity, we assume that there is at least one path between any pair of nodes.
The most basic class of algorithms that can be used to solve the \texttt{Path}$(\startnode,\targetnode)$ problem is the class of \emph{graph search algorithms}.
Two of the most prominent examples of graph search algorithms are \emph{Breadth-First-Search} (\BFS) and \emph{Depth-First-Search} (\emph{DFS}).

Both algorithms start at the starting node $\startnode$ and traverse  the graph iteratively by following its edges.
Intuitively, \emph{DFS}$(\startnode)$ tries to follow a single path starting in $\startnode$ for as long as possible, only returning to a previously considered node and starting a \enquote{new} path if it is strictly necessary.
In contrast to this, \BFS$(\startnode)$ tries to always visit a node as close as possible to the starting node $\startnode$ next.
A visualization of the results of these two algorithms applied to the same graph starting at node $a$ is given in \Cref{fig:AlgorithmExample}.
By remembering which nodes are already completely explored, both of these algorithms find all nodes reachable from the starting node $\startnode$.
}

\redact{
\begin{figure}[H]
    \centering
    \begin{tikzpicture}
    \tikzset{every node/.style={draw, circle, inner sep=2pt, scale=0.66, minimum size=0.8 cm}}
    \def\x{0.075\textwidth}
    \def\y{0.085\textwidth}
    \node (v1) at (-0.1*\x,-0.1*\y) {$a$};
    \node (v2) at (0.9*\x,-0.2*\y) {$b$};
    \node (v3) at (2.1*\x,-0.3*\y) {$c$};
    \node (v4) at (3.2*\x,-0.2*\y) {$d$};
    \node (v5) at (4.3*\x,0) {$e$};
    \node (v6) at (-0.5*\x,\y) {$f$};
    \node (v7) at (\x,\y) {$g$};
    \node (v8) at (2*\x,0.6*\y) {$h$};
    \node (v9) at (2.9*\x,\y) {$i$};
    \node (v10) at (4.2*\x,\y) {$j$};
    \node (v11) at (-0.2*\x,1.9*\y) {$k$};
    \node (v12) at (1.1*\x,2*\y) {$l$};
    \node (v13) at (1.9*\x,2.3*\y) {$m$};
    \node (v14) at (3*\x,2.15*\y) {$n$};
    \node (v15) at (4.15*\x,2.3*\y) {$o$};
    
    \foreach \i/\j in {1/2, 1/6, 1/7, 1/12, 2/7,2/8,2/3,3/7,3/8,3/9,3/4,4/4,4/15,5/10,5/8,6/11,6/14, 7/11,7/8,7/9,7/15,8/13,9/12,9/14,9/15,10/15,11/12,12/13,13/14,14/15}{
        \draw (v\i)--(v\j);
    }
    
    \foreach \i/\j in {1/6,1/12,1/7,1/2, 2/8,2/3,7/11,7/9,6/14,3/4,3/9,8/5,8/13,9/15,5/10}{
        \draw[<-,red, very thick] (v\i)--(v\j);
    }
    
    \begin{scope}[xshift=0.5\textwidth]
    \node (v1) at (-0.1*\x,-0.1*\y) {$a$};
    \node (v2) at (0.9*\x,-0.2*\y) {$b$};
    \node (v3) at (2.1*\x,-0.3*\y) {$c$};
    \node (v4) at (3.2*\x,-0.2*\y) {$d$};
    \node (v5) at (4.3*\x,0) {$e$};
    \node (v6) at (-0.5*\x,\y) {$f$};
    \node (v7) at (\x,\y) {$g$};
    \node (v8) at (2*\x,0.6*\y) {$h$};
    \node (v9) at (2.9*\x,\y) {$i$};
    \node (v10) at (4.2*\x,\y) {$j$};
    \node (v11) at (-0.2*\x,1.9*\y) {$k$};
    \node (v12) at (1.1*\x,2*\y) {$l$};
    \node (v13) at (1.9*\x,2.3*\y) {$m$};
    \node (v14) at (3*\x,2.15*\y) {$n$};
    \node (v15) at (4.15*\x,2.3*\y) {$o$};
    
    \foreach \i/\j in {1/2, 1/6, 1/7, 1/12, 2/7,2/8,2/3,3/7,3/8,3/9,3/4,4/4,4/15,5/10,5/8,6/11,6/14, 7/11,7/8,7/9,7/15,8/13,9/12,9/14,9/15,10/15,11/12,12/13,13/14,14/15}{
        \draw (v\i)--(v\j);
    }
    
    \foreach \i/\j in {1/2,2/8,8/13,13/14,14/15,15/10,10/5,15/4,4/3,3/9,9/7,7/11,11/12,14/6}{
        \draw[<-,blue, very thick] (v\i)--(v\j);
    }
    
    \end{scope}
    
    \end{tikzpicture}
    \caption{Exemplary result of \BFS$(a)$ (left) and \emph{DFS}$(a)$ (right).
    Each node but $a$ points to its parent node.} \label{fig:AlgorithmExample}
\end{figure}
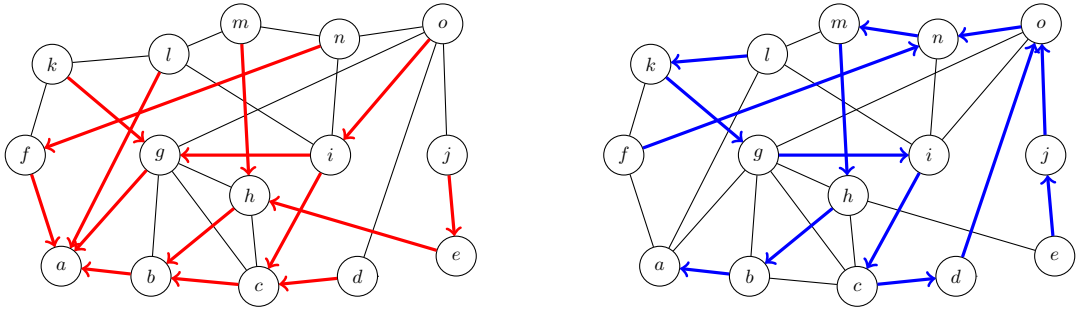
}

\redact{
In particular, the algorithms are typically implemented in a way that the traversed paths can easily be recovered from the data produced by the algorithms.
As both algorithms are very similar, they can be implemented as a realization of a general scheme for finding paths in a graph.
This scheme is given in \Cref{alg:GenericGraphTraversal}.
It uses a generic data structure $D$ that only has to allow for the two basic operations of inserting in and removing nodes from it.
In each step, the algorithm extracts a node $u$ from $D$ and checks for unvisited nodes among all nodes which have an edge towards $u$.
For each such node $w$, the algorithm inserts $w$ into the data structure $D$ and remembers that the node $w$ was reached from $u$ by marking $u$ as the parent of $w$.
To avoid visiting vertices more than once, the node $w$ is then also marked as visited.
After performing this step for each such node, the node $u$ is completely explored and it is not necessary to consider it again.

Depending on the specific data structure that is chosen for $D$, this then yields either the \BFS or the \emph{DFS} algorithm.
More precisely, if $D$ is chosen as a queue that inserts and removes nodes \emph{first-in-first-out}, then \Cref{alg:GenericGraphTraversal} yields the \BFS algorithm.
If $D$ is chosen as a stack that inserts and removes nodes \emph{last-in-first-out}, then one obtains the \emph{DFS} algorithm.
}

\redact{
\begin{algorithm}[H]
\SetAlgoLined
\DontPrintSemicolon
\KwData{graph $G=(\V,\E)$. node $\startnode\in\V$}
\KwResult{calculated parent $p(w)$ for each $w\in\V$}
$D\coloneqq (\startnode)$\;
$p(\startnode)\coloneqq \startnode$\;
Mark $\startnode$ as visited\;
\While{$D\neq\emptyset$}{
$u\coloneqq \texttt{ExtractElement} (D)$\;
%Remove $u$ from $D$\;
\ForEach{$w$ that has an edge to $u$}{
    \If{$w$ is not marked as visited}{
        Insert $w$ into $D$\;
        $p(w)\coloneqq u$\;
        Mark $w$ as visited\;
    }
}
}
\Return list of parents $p$
\caption{The generic graph traversal algorithm.
Choosing a queue for $D$ yields the \BFS algorithm, choosing a stack yields the \emph{DFS} algorithm.} \label{alg:GenericGraphTraversal}
\end{algorithm}
}

\redact{
Both variants of this algorithmic scheme can solve the \texttt{Path}$(\startnode,\targetnode)$ problem.
However, as mentioned in \Cref{sec:model}, we want to find a \emph{short} path from $\startnode$ to $\targetnode$.
This is guaranteed if we use the \BFS$(\startnode)$ algorithm as this algorithm always finds shortest paths with respect to the number of edges.
We later argue why this result implies that we are able to find short paths in the neural network representing the manifold of stimuli, even if we cannot guarantee that they are shortest paths.

We now discuss why it is not biologically plausible that graph traversal problems in the brain are solved by exactly one of these algorithms.
The main obstacle is that \Cref{alg:GenericGraphTraversal} requires the data structure $D$ to organize the nodes that still have to be considered, as well as a mechanism to remember which nodes have already been visited.
Especially the data structure $D$ which might have to store a large number of nodes and is in some sense \enquote{global} cannot be implemented in the brain in a way it can be implemented in a computer.
The reason is that individual neurons in a neural network can only access local information or information that was just sent to them by a pre-synaptic neuron.
In a neural network, however, neurons are only able to communicate with their synaptic neighbors via sending and receiving electric current.

As discussed in \Cref{sec:Relation_To_Existing_Algorithms}, our network configuration yields a wave propagation that behaves like a \enquote{parallelized} version of \BFS where a set of nodes can be visited simultaneously.
This also explains how using a wave propagation algorithm can find short paths, but not necessarily shortest paths:
A neuron potentially receives current from more than one neuron, hence it is not possible to uniquely retrace the path to the starting node.
However, as wave propagation behaves like a parallelized \BFS algorithm, the paths that can be obtained via backtracking will never be too long.
Although this behavior has some similarities with other well-understood graph problems like virus propagation \cite{Bonnet:2017,Kephart:1991,VanMieghem:2009} or diffusion processes \cite{Ibe:2013} in networks, the respective theories are not directly applicable to our specific scenario.
}

% =====================================================================
\subsection*{Neuronal Network Setup -- Exemplary Implementation of the Model}\label{sec:methods:implementation}
% =====================================================================

% =====================================================================
\paragraph{Splitting Dynamics to Two Network Layers}

As described in \Cref{sec:model:dynamics}, for our numerical implementation of the model, we separated the two different types of dynamics into distinct layers of neurons, the \emph{continuous attractor layer} and the \emph{wave propagation layer}.
The split into two layers makes the model more transparent and ensures that parameter changes have limited and traceable effects on the over-all dynamics. 
As an additional simplification, we do not explicitly model the feed-forward connections which drive the wave propagation layer, but we rather directly activate certain neurons in this layer.

%As a second option, we believe that in principle both types of dynamics could appear in a single layer which combines both the continuous attractors and the propagation of wave fronts:
%For example, if the inhibition between neurons were subject to a time delay it would still have the full effect on a stationary peak of activity, but at the same time it might allow propagating wave fronts to travel through an area of the network before inhibition kicks in.
%We have not implemented such a single-layer network yet, but we think that it would add further to the biological plausibility of the model, since it seems more natural for such a network to grow organically.
%We discuss this point in more detail below in \Cref{sec:limitations}.

Activation in the \emph{continuous attractor layer} $C$ represents the start node \startnode, that in the course of the simulation will move towards the target node \targetnode, which is permanently stimulated in the \emph{wave propagation layer} $P$.
Waves of activation are travelling from \targetnode across $P$.
As soon as the wave front reaches a node in $P$ that is connected to a node in proximity to the current activation in $C$, the activation in $C$ is moved towards it.
Thus, every arriving wave front will pull the activation in $C$ closer to \targetnode, forcing the activation to trace back the wave propagation to its origin \targetnode.
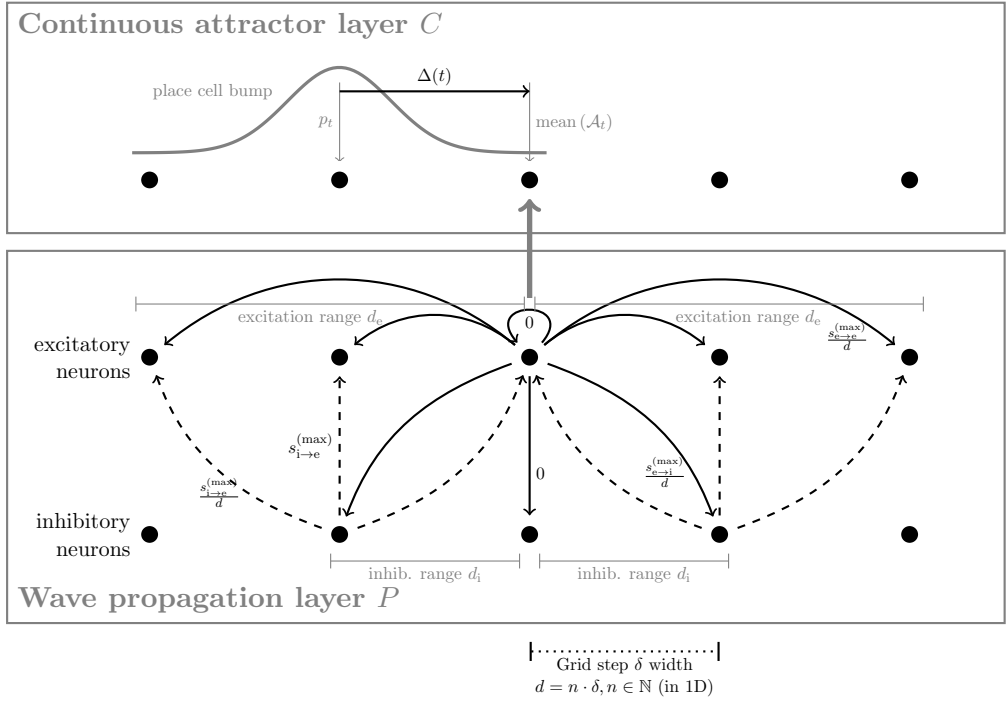
\begin{figure}[tb]
    \centering

        \begin{tikzpicture}
        \tikzset{excitatory/.style={line width=4pt, draw=white, fill=black, circle}}
        \tikzset{inhibitory/.style={line width=4pt, draw=white, fill=black, circle}}
        \tikzset{continuous/.style={line width=4pt, draw=white, fill=black, circle}}
        
        \def\x{2.5}
        \def\y{2.35}
        \def\s{0.6}
        
        \begin{scope}[xshift = 0.69*\x cm, yshift=3.*\y]
        \begin{axis}[
            no markers, domain=0:8, samples=100,
            every axis y label/.style={at=(current axis.above origin),anchor=south},
            every axis x label/.style={at=(current axis.right of origin),anchor=west},
            height=1.25*\y cm, width=3.25*\x cm,
            hide axis
            ]
            \addplot [very thick,gray] {gauss(4,1)};
            \end{axis}
        \end{scope}
        
        \node[scale=\s, gray] at (1.333*\x,0.5*\y) {place cell bump}; 
        
        \draw[->, gray] (2*\x, 0.55*\y)--node[left, scale=\s] {$p_t$} ++(0,-0.45*\y);
        \draw[->, gray] (3*\x, 0.55*\y)--node[right, scale=\s] {$\mean{\At}$} ++(0,-0.45*\y);
        \draw[->, thick, black] (2*\x, .5*\y)--node[above, scale=\s] {$\Delta(t)$}++(\x,0);

        %Labels
        \draw[thick, gray] (0.25*\x,-0.3*\y) rectangle (5.5*\x, 1*\y);
        \node[gray,align=left, anchor=north west] at (0.25*\x, 1*\y) {\textbf{Continuous attractor layer $C$}};
        %Labels
        \draw[thick, gray] (0.25*\x,-2.5*\y) rectangle (5.5*\x, -.4*\y);
        \node[gray, align=left, anchor=south west] at (0.25*\x, -2.5*\y) {\textbf{Wave propagation layer $P$}};
        \node[black, align=right, text width=3cm, scale=1.25*\s] at (0.45*\x,-2*\y) {inhibitory neurons};
        \node[black, align=right, text width=3cm, scale=1.25*\s] at (0.45*\x,-1*\y) {excitatory neurons};

        \foreach \i in {1,...,5}{
            \node[continuous] (c\i) at (\i*\x,0) {};
            \node[excitatory] (e\i) at (\i*\x,-\y) {};
            \node[inhibitory] (i\i) at (\i*\x,-2*\y) {};
        }

        %Drawing in the excitatory layer
        \draw[line width=2pt, black!50, ->] (3*\x,-0.666*\y)--(c3);
        \draw[thick,->] (e3) to[out=50,in=130, looseness=6] node[below, scale=\s] {$0$} (e3);
        \draw[thick,->] (e3) to[out=40, in=140] (e4);
        \draw[thick,->] (e3) to[out=40, in=140] node[below right=0.5cm, scale=\s, pos=0.7] {$\frac{s_\ee\ma}{d}$} (e5);
        \draw[thick,->] (e3) to[out=140, in=40] (e1);
        \draw[thick,->] (e3) to[out=140, in=40] (e2);

        %Connection excitatory/inhibitory layer
        \draw [thick,->] (e3) -- node[pos=0.7, scale=\s, right] {$0$} (i3);
        \draw [thick,->] (e3) to[out=200,in=70, right] (i2);
        \draw [thick,->] (e3) to[out=-20, in=110, left] node[pos=0.8, scale=\s] {$\frac{s_\ei\ma}{d}$} (i4);

        %Connection inhibitory/excitatory layer
        \draw[thick,->, dashed] (i2) to[out=160, in=290] node[below, scale=\s] {$\frac{s_\ie\ma}{d}$} (e1);
        \draw[thick,->, dashed] (i2)--node[left, scale=\s] {$s_\ie\ma$} (e2);
        \draw[thick,->, dashed] (i2) to[out=20, in=250] (e3);
        \draw[thick,->, dashed] (i4) to[out=160, in=290] (e3);
        \draw[thick,->, dashed] (i4)-- (e4);
        \draw[thick,->, dashed] (i4) to[out=20, in=250] (e5);

        %Misc
        \draw[|-|, gray] (3.05*\x, -2.15*\y)--node[below, scale=\s, gray] {inhib. range $d_\mathrm{i}$} (4.05*\x,-2.15*\y);
        \draw[|-|, gray] (1.95*\x, -2.15*\y)--node[below, scale=\s, gray] {inhib. range $d_\mathrm{i}$} (2.95*\x,-2.15*\y);
        \draw[|-|, gray] (3.025*\x, -0.7*\y)--node[below, scale=\s, gray, pos=0.55] {excitation range $d_\mathrm{e}$} (5.075*\x,-0.7*\y);
        \draw[|-|, gray] (0.925*\x, -0.7*\y)--node[below, scale=\s, gray, pos=0.45] {excitation range $d_\mathrm{e}$} (2.975*\x,-0.7*\y);

        \draw[thick, dotted, |-|] (3*\x, -2.66*\y)--node[below, scale=\s, text width=4cm, align=center] {Grid step $\delta$ width $d=n\cdot\delta, n\in\mathbb{N}$ (in 1D)} (4*\x, -2.66*\y);

        \end{tikzpicture}

    \caption{%
        Connectivity of the neurons.
        For simplicity, this visualization only contains a 1D representation.
        In the wave propagation layer, excitatory synapses are drawn as solid arrows, dashed arrows indicate inhibitory synapses.
        Upon its activation, the central excitatory neuron stimulates a ring of inhibitory neurons that in turn suppress circles of excitatory neurons to prevent an avalanche of activation and support a circular wave-like expansion of the activation across the sheet of excitatory neurons.
        Furthermore, overlap between the active neurons in $C$ and $P$ is used to compute the direction vector $\Delta(t)$ used for biasing synapses in $C$ and thus shifting activity there.
        %
        %After moving the continuous attractor activity, coupling between the layers is deactivated for a while.
        %This ensures that the place cell activation is moving towards the incoming activation wave front instead of following it after it passed through it.
        }
    \label{fig:LayerConnectivity}
\end{figure}

In detail, these dynamics require a very specific network configuration which is described in the following.
\Cref{fig:LayerConnectivity} contains a general overview of the intra- and inter-layer connectivity used in the model and our simulations.

% =====================================================================
%\subsubsection*{Wave Propagation Layer} \label{sec:methods:implementation:wave_propagation_layer}

\paragraph{Spiking Neuron Model in the Wave Propagation Layer}
In the performed experiments, the wave propagation layer $P$ is constructed with an identical number of excitatory and inhibitory Izhikevich neurons \cite{Izhikevich:2003, Izhikevich:2004}, that cover a regular quadratic grid of $41\times 41$ points on the manifold of stimuli.
\redact{
The spiking behavior of each artificial neuron is modeled as a function of its membrane potential dynamics $v(t)$ using the two coupled ordinary differential equations
\begin{align}
  \frac{\mathrm{d}}{\mathrm{d}t}v &= 0.04 v^2 +5 v + 140 - u + I, \label{eq:izhikevic:v}\\
  \frac{\mathrm{d}}{\mathrm{d}t}u &= a\cdot(b v-u).
  \intertext{Here, $v$ is the membrane potential in \si{\milli\volt}, $u$ an internal recovery variable, and $I$ represents synaptic or DC input current. 
  The internal parameters $a$ (scale of $u$ / recovery speed) and $b$ (sensitivity of $u$ to fluctuations in $v$) are dimensionless.
  Time $t$ is measured in \si{\milli\second}.
  If the membrane potential grows beyond the threshold parameter $v\geq\SI{30}{\milli\volt}$, the neuron is spiking and the variables are reset as follows.}
  v &\leftarrow c,\\
  u &\leftarrow u+d.
\end{align}
Again, $c$ (after-spike reset value of $v$) and $d$ (after-spike offset value of $u$) are dimensionless internal parameters.
}

\begin{table}[tb]
    \begin{subtable}[b]{0.3\textwidth}
        \centering
        \begin{tabular}{rSS}
          \toprule
              & \multicolumn{1}{c}{excitatory}
              & \multicolumn{1}{c}{inhibitory} \\
              & \multicolumn{1}{c}{RS}
              & \multicolumn{1}{c}{FS}\\
          \midrule
          $a$ & 0.02 & 0.1 \\
          $b$  & 0.2  & 0.2 \\
          $c$ & -65  & -65 \\
          $d$ & 8    & 2   \\
          \bottomrule
        \end{tabular}
        \caption{Neuron model parameters (homogeneous setup).}
        \label{tab:params_P:neurons}
    \end{subtable}
    \hfill
    \begin{subtable}[b]{0.33\textwidth}
        \centering
        \begin{tabular}{rcc}
          \toprule
              & \multicolumn{1}{c}{excitatory}
              & \multicolumn{1}{c}{inhibitory} \\
              & \multicolumn{1}{c}{RS \ldots CH}
              & \multicolumn{1}{c}{LTS \ldots FS}\\
          \midrule
          $a$ & 0.02             & $0.02 + 0.08r_i$ \\
          $b$  & 0.2             & $0.25 - 0.05r_i$ \\
          $c$ & $-65 + 15 r_e^2$ & -65 \\
          $d$ & $8 - 6 r_e^2$    & 2 \\
          \bottomrule
        \end{tabular}
        \caption{Neuron model parameters (heterogeneous setup).}
        \label{tab:params_P:heterogeneous}
    \end{subtable}
    \hfill
    \begin{subtable}[b]{0.225\textwidth}
        \centering
        \renewcommand{\arraystretch}{1.2}
        \begin{tabular}{rS}
          \toprule
          $s_\ee\ma$  & \SI{50}{\phantom{\seema}} \\
          $s_\ei\ma$  & \SI{0.5}{\seema} \\
          $s_\ie\ma$  & \SI{-9}{\seema} \\
          \midrule
          $d_\mathrm{e}$ & 2 \\
          \bottomrule
        \end{tabular}
        \caption{Synaptic strength parameters.}
        \label{tab:params_P:connectivity}
    \end{subtable}
    \caption{Parameters used in our simulations of the wave propagation layer $P$.}
    \label{tab:params_P}
\end{table}

If not stated otherwise in the following, the parameters listed in \Cref{tab:params_P:neurons} were used for the \redact[Izhikevich neurons]{spiking neuron model} in $P$.
They correspond to regular spiking (RS) excitatory and fast spiking (FS) inhibitory neurons.
In contrast to \cite{Izhikevich:2003}, neuron properties were not randomized to allow for reproducible analyses.
The effect of a more biologically plausible heterogeneous neuron property and synaptic strength distribution is analyzed in \Cref{sec:methods:experiments}.
Compared to \cite{Izhikevich:2003}, the coupling strength in $P$ is large to account for the extremely sparse adjacency matrix as every neuron is only connected to its few proximal neighbours in our configuration.
Whenever a neuron in $P$ is to be stimulated externally, a DC current of $I=25$ is applied to it\redact{, \cf \Cref{eq:izhikevic:v}}.
As in \cite{Izhikevich:2003}, the simulation time step was fixed to \SI{1}{\milli\second} with one sub-step in $P$ for numerical stability.

\paragraph{Synaptic Connections in the Wave Propagation Layer}
\redact{
    As described before, neurons in $P$ correspond to reachable locations in the manifold of stimuli.
    Thus, it is plausible to assume that neurons representing near-by locations in a suitable metric on the respective manifold will also be closely connected.
    Assuming that neurons will not have a very strictly defined region of responsibility, but there will also be some overlap, this is consistent with a Hebbian learning approach:
    Neurons that are sensitive to nearby regions will often fire at the same instant in time, strengthening their mutual connectivity.
}

As depicted in \Cref{fig:LayerConnectivity}, the excitatory neurons are driving nearby excitatory and inhibitory neurons with a synaptic strength of
\begin{align}
    s_\ee(d) &\coloneqq
        \begin{cases}
            \dfrac{s_\ee\ma}{d}, & \text{for } 0 < d\leq d_\mathrm{e} \\[.75em]
            \hfill 0 , & \text{else}
        \end{cases}\label{eq:see},
\end{align}
where $s_\ei(d)$ is defined analogously.
Here, $d$ is the distance between nodes in the manifold of stimuli.
For simplicity, we model this manifold as a two-dimensional quadratic mesh with grid spacing $\delta=1$ where some connections might be missing.
The choice $s\propto\nicefrac{1}{d}$ was made to represent the assumption that recurrent coupling will be strongest to nearest neighbours and will decay with distance.
Note that \eqref{eq:see} in particular implies that we have $s_\ee(0),s_\ei(0)=0$, which prevents self-excitation.
To restrict to only localized interaction, we exclude interaction beyond a predefined excitation range $d_\mathrm{e}$ and inhibition range $d_\mathrm{i}$, respectively.
Values of the parameters in the expressions for the synaptic strengths used in the simulations are given in \Cref{tab:params_P:connectivity}.

The inhibitory neurons suppress activation of the excitatory neurons by reducing their input current via synaptic strength
\begin{align}
    s_\ie(d) &\coloneqq
        \begin{cases}
            s_\ie\ma, & \text{for } d = 0 \\[.75em]
            \dfrac{s_\ie\ma}{d}, & \text{for } 0 < d\leq d_\mathrm{s} \\[.75em]
            \hfill 0 , & \text{else}
        \end{cases}\quad.
\end{align}

\begin{figure}[tb]
    \centering
    \includegraphics[width=0.66\textwidth]{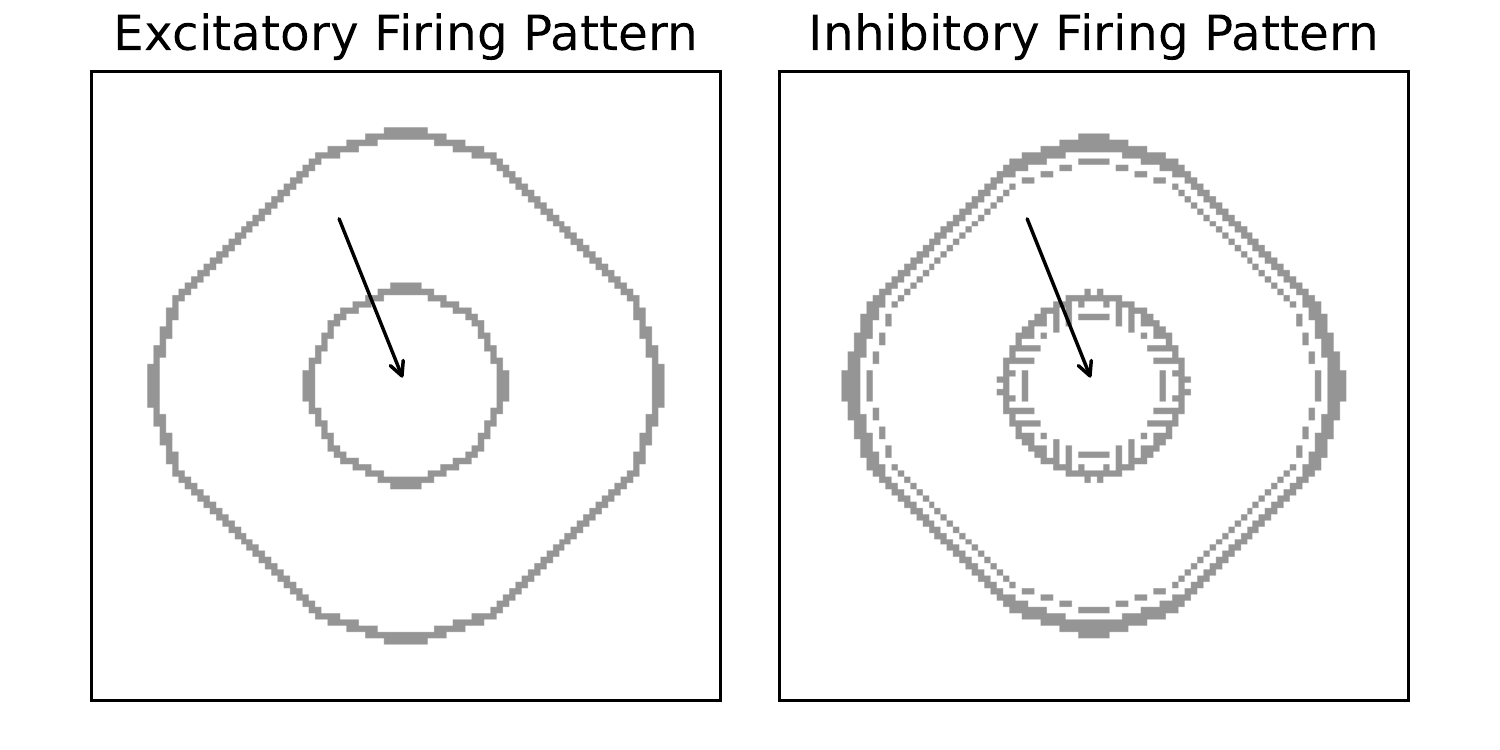}
    \caption{%
        Activity patterns of the excitatory and inhibitory neurons on a $101\times101$ quadratic neuron grid.
        Spiking neurons are shown as gray areas.
        One excitatory neuron at the grid center (arrow) is driven by an external DC current to regular spiking activity.
        Due to the nearest-neighbour connections, this activity is propagating in patterns that resemble a circular wave structure.
        The inhibitory neurons prevent catastrophic avalanche-like dynamics by suppressing highly active regions.
        \redact{The specific pattern shape is an artifact of the underlying regular grid structure and thus not perfectly circular.
        This could be alleviated using, \eg a hexagonal instead of a quadratic mesh of neurons.}
        }
    \label{fig:ExcitatoryWaves}
\end{figure}

\paragraph{Wave Propagation Dynamics}
The described setup allows for wave-like expansion of neuronal activity from an externally driven excitatory neuron as shown in \Cref{fig:ExcitatoryWaves}.

\redact{
    If the activity of the excitatory neurons grows too much in a region, the respective inhibitory neurons will start spiking to eventually suppress activity locally.
    This suppression happens with a delay of two time steps due to the causal signal travelling time through $s_\ee$ and $s_\ei$, but could also be implemented via different synaptic time constants, \ite AMPA (excitatory) vs. GABA\,A (inhibitory).
    Thus, the inhibitory neurons prevent an avalanche-like activity by turning off active excitatory neurons.

    As can be seen in \Cref{fig:ExcitatoryWaves}, this effectively means that propagating signals in the excitatory sub-network are followed by similarly shaped propagating signals in the inhibitory sub-network.
    In this respect, signal propagation does not behave like physical waves, such as ripples on water:
    They do not interfere in constructive and destructive manner to form interference patterns.
    Instead, activity stops where to propagating signals touch as shown in \Cref{fig:annihilation}.
    This is an important property in our setup as it ensures that signals do not run through each other in the wave propagation layer but do mutually annihilate.
    Thus, the wave fronts tend to form stable and continuous patterns and activation of the continuous attractor layer from different directions is vastly reduced.
}
\redact{
    \begin{figure}[H]
        \hfill%
        \includegraphics[width=0.25\textwidth]{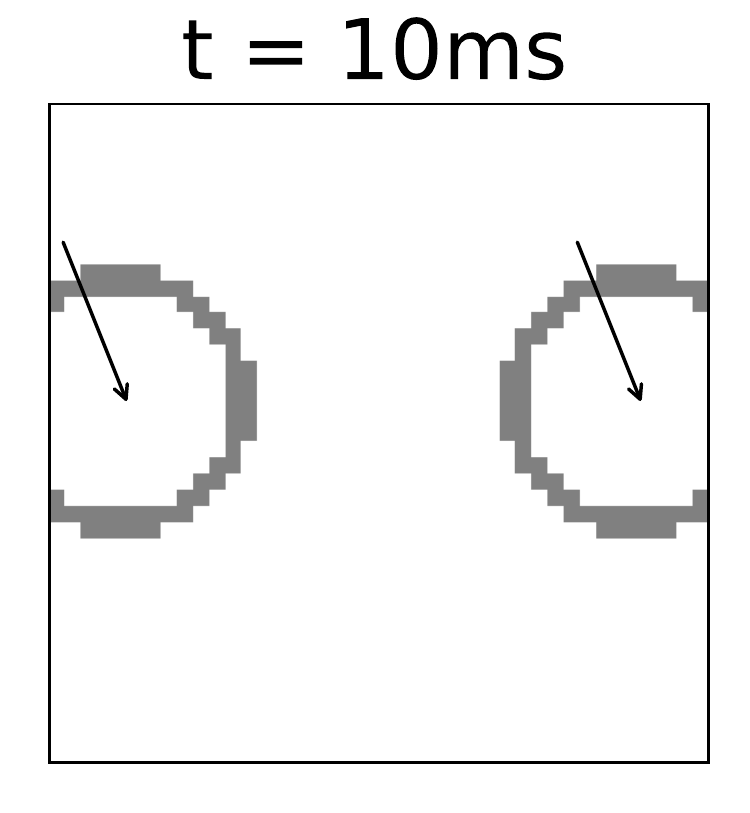}\hfill%
        \includegraphics[width=0.25\textwidth]{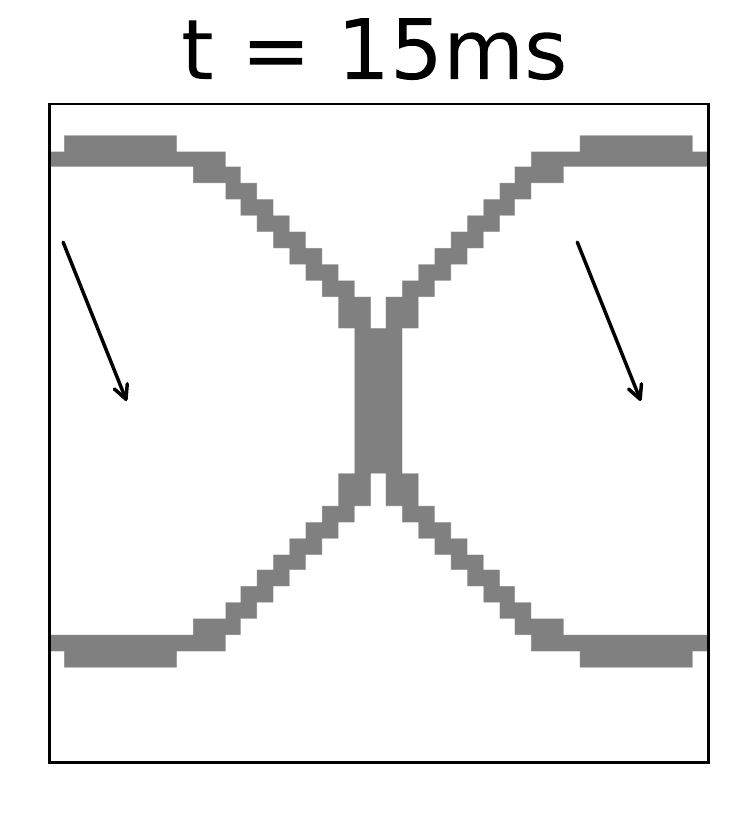}\hfill%
        \includegraphics[width=0.25\textwidth]{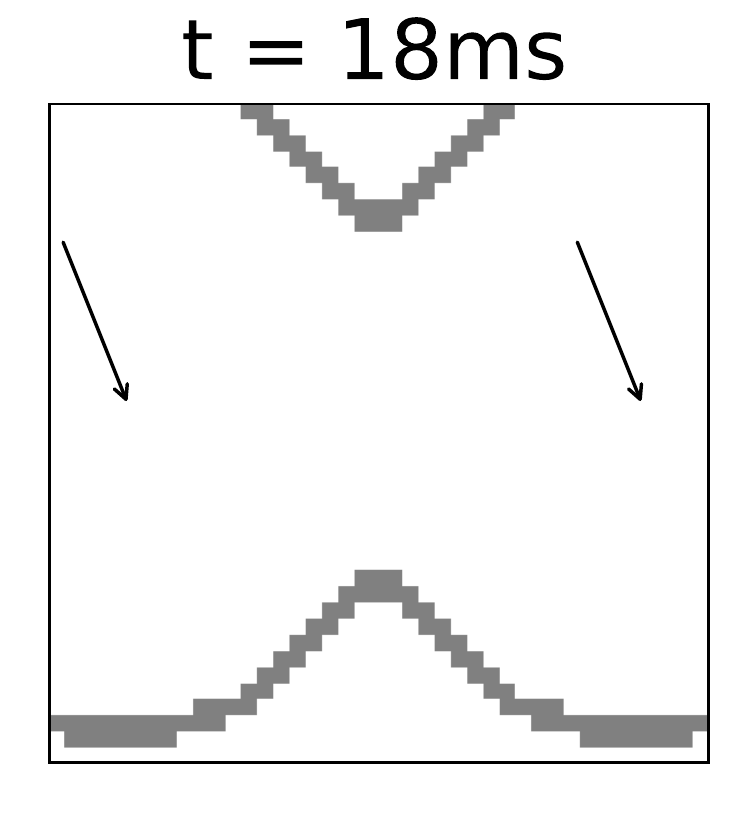}\hfill%
        \caption{%
            Activity patterns of the excitatory neuron grid where two neurons are driven to periodic spiking activity (arrows) at different instants in time.
            Again, spiking neurons are shown as gray areas and neuronal connections are set up as described in \Cref{sec:methods:implementation}
            As soon as the signal propagation fronts touch, they annihilate each other due to the inhibitory activity that accompanies them.
            Instead of forming interference patterns or travelling through each other, the remaining wave fronts merge and continue propagating as a well-defined line of activity.
            }
        \label{fig:annihilation}
    \end{figure}
}

With the capability of propagating signals as circular waves from the target neuron \targetnode across the manifold of stimuli in~$P$, it is now necessary to set up a representation of the start neuron~\startnode in~$C$.
This will be done in the following subsection before the coupling between~$P$ and~$C$ will be described.

% =====================================================================
%\subsubsection*{Continuous Attractor Layer} \label{sec:methods:implementation:continuous_attractor_layer}

\paragraph{Neuron Model for Place Cell Dynamics}
The \emph{continuous attractor layer} $C$, implements a sheet of neurons that models the functionality of a network of place cells in the human hippocampus using rate-coding neurons \cite{Okeefe1971, Okeefe1976} and thus the manifold of stimuli.
As for the wave propagation layer, we also use a quadratic $41\times 41$ grid of neurons for this layer.
Activation in the continuous attractor layer will appear as bump, the center of which represents the most likely current location on the manifold of stimuli.

This bump of activation is used to represent the current position in the graph of synaptic connections representing the cognitive map.
Planning in the manifold of stimuli thus amounts to moving the bump through the sheet of neurons where each neuron can be thought of as one node in this graph.
With respect \eg to the robot arm example in \Cref{fig:examples:robot}, the place cell bump represents the current state of the system \ite the current angles of the arm's two degrees-of-freedom. As the bump moves through the continuous attractor layer, and thus through the graph, the robot arm will alter its configuration creating a movement trajectory through the 2D space.

\paragraph{Synaptic Connectivity to Realize Continuous Attractor Dynamics}
Our methodology for modelling the continuous attractor place cell dynamics adapts the computational approach used in \cite{Guanella2007} by including a computational consideration for synaptic connections between continuous attractor neurons and an associated update rule that depends on information from the wave propagation layer $P$.

The synaptic weight function connecting each neuron in the continuous attractor sheet to each other neuron is given by a weighted Gaussian.
This allows for the degrading activation of cells in the immediate neighbourhood of a given neuron and the simultaneous inhibition of neurons that are further away, thus giving rise to the bump-shaped activity in the sheet itself.
The mathematical implementation of these synaptic connections also allows for the locus of activation in the sheet to be shifted in a given direction which is, in turn, how the graph implemented by this neuron sheet is able to be traversed.

The synaptic weight $w_{\Vec{i},\Vec{j}}\in\mathbb{R}^{(N_x\times N_y)\times(N_x\times N_y)}$ connecting a neuron at position $\Vec{i}=(i_x, i_y)$ to a neuron at position $\Vec{j}=(j_x, j_y)$ is given by
\begin{align}
    w_{\Vec{i},\Vec{j}} &\coloneqq J\cdot\exp\left(-\frac{1}{\sigma^{2}}\left\lVert\left(\frac{i_x-j_x}{N_x},\frac{i_y-j_y}{N_y}\right)+\Vec{\Delta}(t)\right\rVert^{2}\right)-T\,.\label{eq:first_weight_function}
\end{align}
Here, $J$ determines the strength of the synaptic connections, $\lVert\cdot\rVert$ is the Euclidean norm, $\sigma$ modulates the width of the Gaussian, $T$ shifts the Gaussian by a fixed amount, $\Vec{\Delta}(t)$ is a direction vector which we discuss in detail later, and $N_{x}$ and $N_{y}$ give the size of the two dimensions of the sheet.

In order to update the activation of the continuous attractor neurons and to subsequently move the bump of activation across the neuron sheet, we compute the activation $A_{\Vec{j}}$ of the continuous attractor neuron $\Vec{j}$ at time $t+1$ using
\begin{align}
    B_{\Vec{j}}(t+1) &= \sum_{\Vec{i}}A_{\Vec{i}}(t)w_{\Vec{i},\Vec{j}}(t)\,,\\
    A_{\Vec{j}}(t+1) &= (1-\tau)B_{\Vec{j}}(t+1)+\tau\frac{B_{\Vec{j}}(t+1)}{\sum_{\Vec{i}} A_{\Vec{i}}(t)}\,,
\end{align}
where $B_{\Vec{j}}(t+1)$ is a transfer function that accumulates the incoming current from all neurons to neuron $\Vec{j}$ and $\tau$ is a fixed parameter that determines stabilization towards a floating average activity.

\begin{table}[tb]
    \centering
    \begin{tabular}{rlS}
      \toprule
      $\sigma$ & Gaussian width               & 0.03 \\
      $T$      & Gaussian shift               & 0.05 \\
      $J$      & Synaptic connection strength & 12   \\
      $\tau$   & Stabilization strength       & 0.8  \\
      \bottomrule
    \end{tabular}
    \caption{Parameters for the continuous attractor layer $C$.}
    \label{tab:params_C}
\end{table}

Simulation parameters for the continuous attractor layer $C$ are given in \Cref{tab:params_C}.
They have been manually tuned to ensure development of stable, Gaussian shaped activity with an effective diameter of approximately twelve neurons in $C$.

As in \cite{Guanella2007}, a direction vector $\Vec{\Delta}(t)\in \mathbb{R}^2$ has been introduced in \Cref{eq:first_weight_function}.
It has the effect of shifting the synaptic weights in a particular direction which in turn causes the location of the activation bump in the attractor layer to shift to a neighbouring neuron.
In other words, it is this direction vector that allows the graph to be traversed by informing the place cell sheet from which direction the wave front is coming in $P$.
Thus all that remains for the completion of the necessary computations is to compute $\Vec{\Delta}(t)$ as a function of the propagating wave and the continuous attractor position. 

% =====================================================================
%\subsubsection*{Layer Interaction Algorithm} \label{sec:methods:implementation:layer_interaction}

\paragraph{Layer Interaction - Direction Vector}
The interaction between the wave propagation layer $P$ and the continuous attractor layer $C$ is mediated via the direction vector $\Vec{\Delta}(t)$.
The direction vector is computed such that it points from the center of the bump of activity towards the center of the overlap between bump and incoming wave as follows.
Let $\mathcal{C}_t$ and $\mathcal{P}_t$ denote the sets of positions of active neurons at time $t$ in layer $C$ and $P$, respectively.
Note that each possible position corresponds to exactly one neuron in the wave propagation layer and exactly one neuron in the continuous attractor layer as they have the same spatial resolution in the implementation.
Now let $\At \coloneqq \mathcal{C}_t\cap\mathcal{P}_t$.
Then,
\begin{align}
    \mean{\At} &= \frac{1}{\left|\At\right|}\sum_{\Vec{i}\in\At}\Vec{i}
    \intertext{%
    is the average position of overlap.
    We compute the direction vector from the current position $p_t$ of the central neuron in the continuous attractor layer activation bump to $\mean{\At}$ via}
     \Vec{\Delta}(t) &= \mean{\At} - p_t\,.
\end{align}
%Now, if we only used $\Vec{\Delta}(t)$ to update the position of the continuous attractor layer activity bump in the subsequent time step while there is still propagating activity on $P$, it would be very likely that $\mathcal{I}_{\mathcal{A}_{t+1}}$ would contain points behind the $p_{t+1}$.
%Thus, $\Vec{\Delta}(t+1)$ would point into the opposite direction, eradicating the progress towards the target node \targetnode from the previous time step by pulling it backwards.

\paragraph{Layer Interaction - Recovery Period}
In order to prevent the wave from interacting with the back side of the bump in $C$, we introduce a recovery period $R$ of a few time steps after moving the bump.
During $R$, which is selected as the ratio of bump size to wave propagation speed, $\mathcal{A}_{t}$ is assumed to be empty, which prevents any further movement.
In our experiments, we used $R=\SI{12}{\milli\second}$.

\redact{
It is worth acknowledging at this point that this approach of connecting the two layers, which we have chosen for reasons of simplicity, is somewhat artificial.
%Both the continuous attractor layer and the wave propagation layer implement models of biological neurons whereas here we compute the direction vector using no neuron model at all. While there is evidence to suggest that neurons are able to encode vectors, particularly in spatial cognition (for a recent review see \cite{Bicanski:2020}, here we have decided to compute the direction vector straightforwardly for the sake of simplicity. 
We discuss this and other limitations of our implementation in \Cref{sec:discussion:merging_layers}.
}

% =====================================================================
\subsection*{Numerical Experiments}\label{sec:methods:experiments}
% =====================================================================

In order to test the complex neuronal network configuration described in \Cref{sec:model,sec:methods:implementation} and to study its properties and dynamics, we performed numerical experiments using multiple different setups.
Source code used for our studies is published at \cite{github_source_code}.
Results of our simulations are presented in \Cref{sec:model:experiments}.
In the following, we will add some more in-depth analyses on specific properties of the model as observed in the simulations.

\paragraph{Transmission Velocity}
In our setup, no synaptic transmission delay, as \eg in \cite{Izhikevich2006}, is implemented.
As, due to the strong nearest-neighbour connectivity, only few pre-synaptic spiking neurons are sufficient to raise the membrane potential above threshold, the waves are travelling across $P$ with a velocity of approximately one neuronal \enquote{ring} per time step, \cf \Cref{fig:simple_setup}.
In contrast, the continuous attractor can only move a distance of at most half its width per incoming wave.
Accordingly, its velocity is tightly coupled to the spike frequency of the stimulated neuron while still being bound due to the recovery period $R$.
\redact{In the specific case of the simulation in \Cref{fig:simple_setup}, in total nine wave fronts were observed to be required traversing the Gaussian continuous attractor activity zone to finally pull it on a straight line to its destination over a distance of $d=45.25$.}

\paragraph{Obstacles and Complex Setups}
In the S-shaped maze \Cref{fig:mazes:s}, the continuous attractor activity moves towards the target node \targetnode on a direct path around the obstacles.
Due to the optimal path being more than two times longer than in \Cref{fig:simple_setup}, the time to reach the target is accordingly longer as well.
This is also in line with the required travel times from \startnode to \targetnode in \Cref{fig:mazes:block,fig:mazes:complex}, where -- despite its complexity -- a path through the maze is found fastest due to it being shorter than in the other cases of \Cref{fig:mazes}.
This observation is also evidenced by the fact that our model is a parallelized version of \BFS, \cf \redact[\Cref{sec:Relation_To_Existing_Algorithms}]{\Cref{sec:Relation_To_Existing_Algorithms,app:graph_traversal_algorithms}}, which is guaranteed to find the shortest path in an unweighted and undirected graph.

\redact{
    \paragraph{Symmetric Paths}
    An additional interesting observation can be made in the central block setup, \Cref{fig:mazes:block}:
    The setup is perfectly symmetric with respect to the two possible paths.
    Thus, in principle it can not be solved with our model.
    However, after interaction with several wave fronts, a minor shift of the continuous attractor position occurs due to numerical instability.
    This is further emphasized by subsequent incoming waves, finally pulling the continuous attractor onto a path to the target node \targetnode.
    While such numerical instabilities are clearly resulting from the specific implementation of our model on a computer system, also organically grown biologic networks will never be perfectly symmetric.
    Here, natural variations in synaptic connectivity and neuron properties will break potential symmetries, favoring one of the possible paths.
    In the following, we inspect the influence of these variations on the overall performance of the model.
}

\paragraph{Heterogeneous Neuron Properties and Synaptic Strengths}
In the simulation experiments described up to now, a homogeneous wave propagation layer $P$ is employed.
There, all neurons are subject to the same internal parameters, being either regular spiking excitatory neurons or fast spiking inhibitory neurons.
Also, synaptic strengths are strictly set as described previously with parameters from \Cref{tab:params_P:connectivity}.
This setup is rather artificial.
Natural neuronal networks will exhibit a broad variability in neuron properties and in the strength of synaptic connectivity.

To account for this natural variability, we randomized the individual neuron's internal properties as suggested in \cite{Izhikevich:2003}, see \Cref{tab:params_P:heterogeneous}.
As in \cite{Izhikevich:2003}, heterogeneity is achieved by randomizing neuron model parameters using random variables $r_e$ and $r_i$ for each excitatory and inhibitory neuron.
These are equally distributed in the interval $[0; 1]$ and vary neuron models between regular spiking ($r_e=0$) and \emph{chattering (CH, $r_e=1$)} or fast spiking ($r_i=1$) for excitatory neurons and \emph{low-threshold spiking (LTS, $r_i=0$)} for inhibitory neurons.
By squaring $r_e$, excitatory neuron distribution is biased towards RS.
In addition, after initializing synaptic strengths in $P$, we randomly varied them individually by up to $\pm\SI{50}{\percent}$.

\begin{figure}[tb]
    \includegraphics[width=0.16\textwidth]{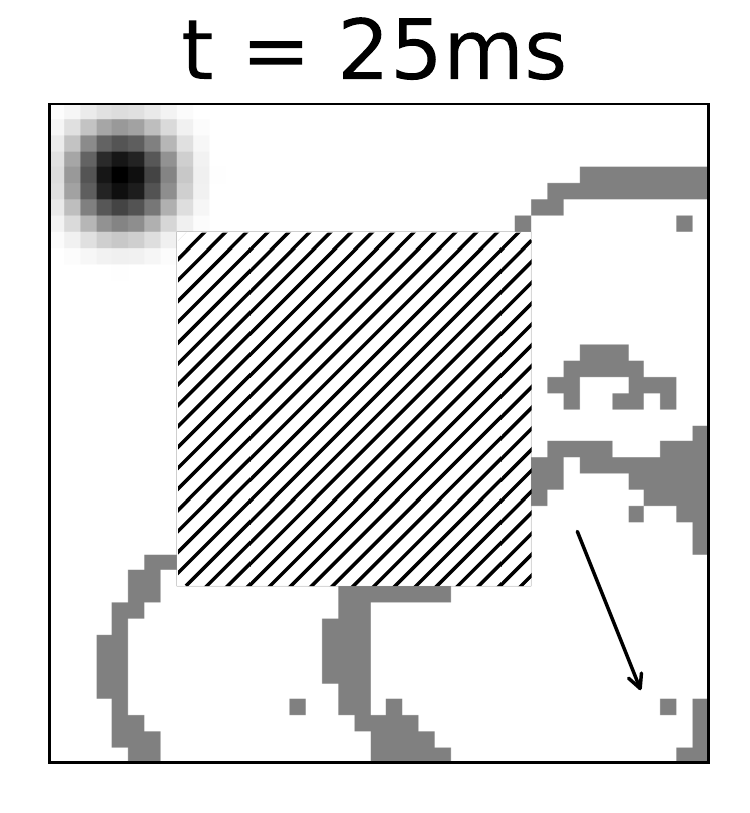}\hfill%
    \includegraphics[width=0.16\textwidth]{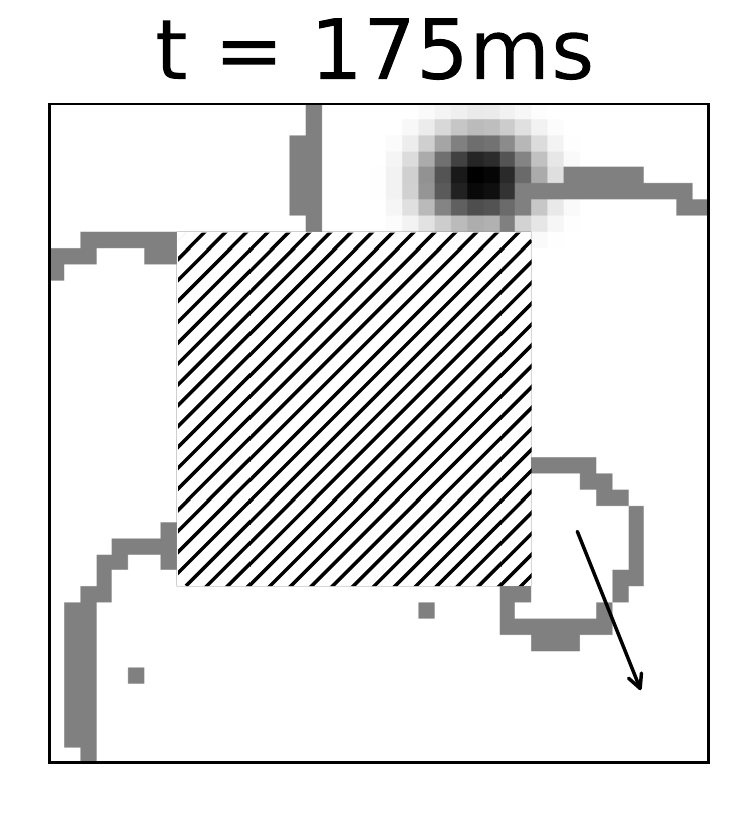}\hfill%
    \includegraphics[width=0.16\textwidth]{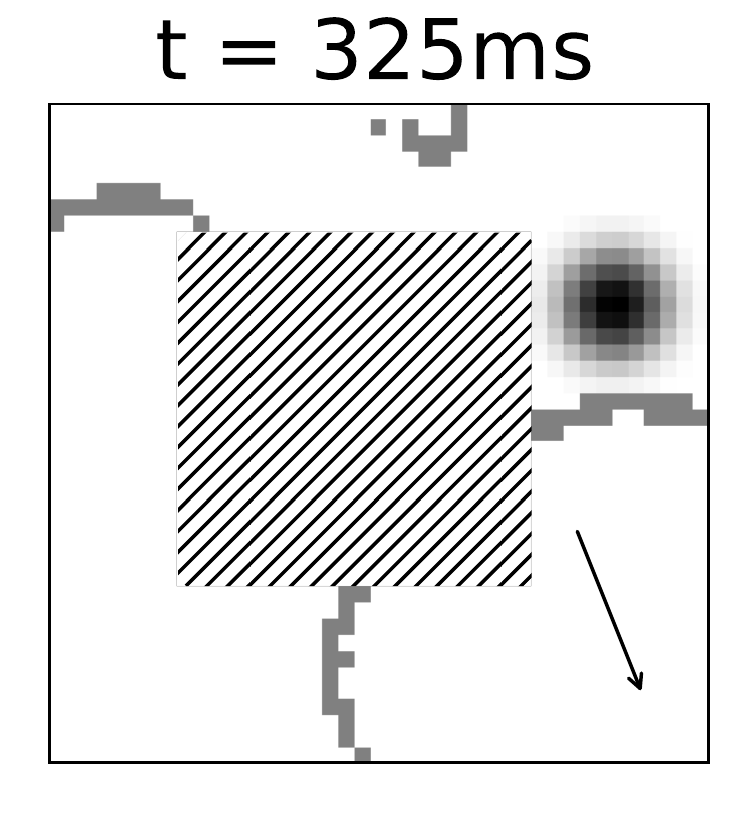}\hfill%
    \includegraphics[width=0.16\textwidth]{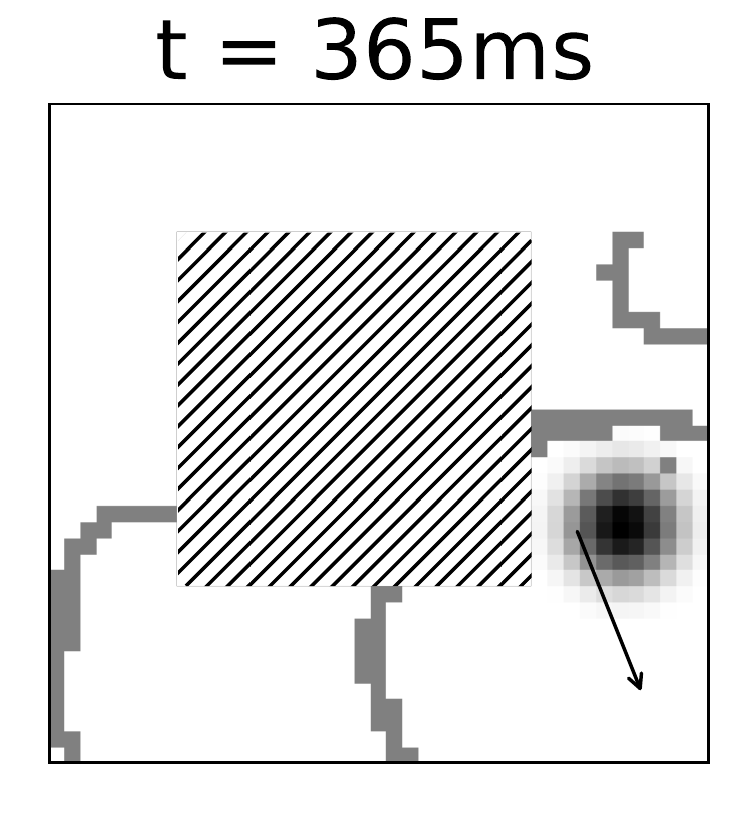}\hfill%
    \includegraphics[width=0.16\textwidth]{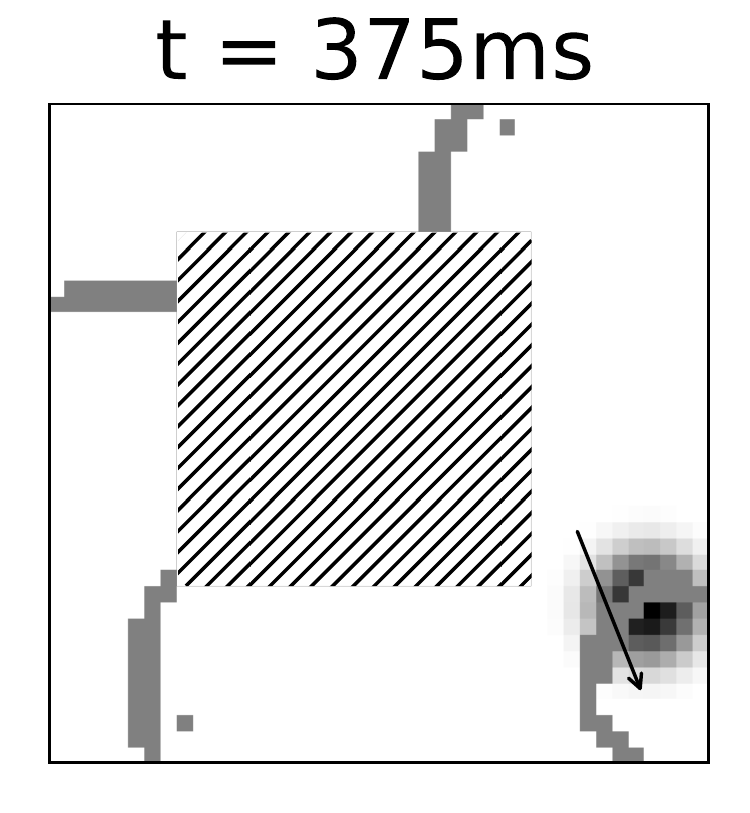}\hfill%
    \includegraphics[width=0.16\textwidth]{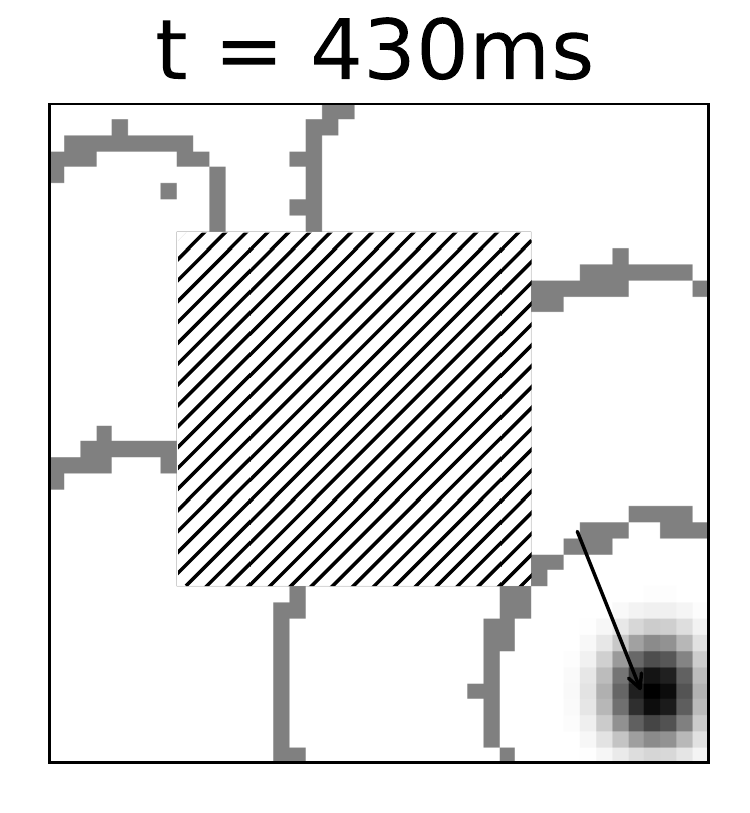}%
    \caption{%
            Block setup as in \Cref{fig:mazes} but with a heterogeneous neuron configuration in $P$.
        }
    \label{fig:mazes:block_heterogeneous}
\end{figure}

Despite this strong modification to the original numerically ideal setup, a structured wave propagation is still possible in $P$ as can be seen in \Cref{fig:mazes:block_heterogeneous}.
While the stereotypical circular form of the wave fronts dissolves in the simulation, they continue to traverse $P$ completely.
As before, they reach the continuous attractor bump and are able to guide it to their origin.
Apparently, the overall connection scheme in $P$ is more important for stable wave propagation than homogeneity in the individual synaptic strengths and neuron properties.

An interesting aspect of this simulation when compared to \Cref{fig:mazes:block} is the apparent capability of solving the graph traversal problem quicker than with the homogeneous neuronal network.
\redact[This]{As already indicated, this} is an artifact of the explicitly broken symmetry in the heterogeneous configuration:
The wave fronts from different directions differ in shape when arriving at the initial position of the continuous attractor layer activity.
Thus, one of them is immediately preferred and target-oriented movement of the bump starts earlier than before.
This capability of breaking symmetries and thus quickly resolving ambiguous situations is an explicit advantage of the more biologically realistic heterogeneous configuration.

\printbibliography

\end{document}